\documentclass[conference]{IEEEtran}
\IEEEoverridecommandlockouts
\pdfoutput=1
\usepackage[T1]{fontenc}
\usepackage[utf8]{inputenc}
\usepackage{cite}
\usepackage{amsmath,amssymb}
\usepackage{graphicx}
\usepackage{booktabs}
\usepackage{array}
\usepackage{textcomp}
\usepackage[hyphens]{url}
\usepackage[hidelinks]{hyperref}
\hypersetup{
  pdftitle={Aqua2CAD: Observation-Guided Point Cloud Correction and Completion for CAD Reconstruction from Underwater Stereo Images},
  pdfauthor={Weizhao Huang, Jiayi Zhang, Pangkit Fong, Chongrong Fang, Jianping He}
}

\makeatletter
\newcommand{\teasercaption}[1]{\def\@captype{figure}\caption{#1}}
\makeatother

\title{\LARGE \bf Aqua2CAD: Observation-Guided Point Cloud\\Correction and Completion for CAD Reconstruction\\from Underwater Stereo Images}

\author{Weizhao Huang$^{1}$, Jiayi Zhang$^{2}$, Pangkit Fong$^{3}$, Chongrong Fang$^{3}$ and Jianping He$^{3*}$%
\thanks{$^{1}$Weizhao Huang is with the School of Integrated Circuits, Shanghai Jiao Tong University, Shanghai, China.
{\tt\small huangweizhao@sjtu.edu.cn}}%
\thanks{$^{2}$Jiayi Zhang is with the School of Mathematical Sciences, University of Nottingham Ningbo China, Ningbo, China.
{\tt\small jiayi.zhang718@gmail.com}}%
\thanks{$^{3}$Pangkit Fong, Chongrong Fang and Jianping He are with the Department of Automation, School of Automation and Perception, Shanghai Jiao Tong University, Shanghai, China.
{\tt\small \{fpjgaoge, crfang, jphe\}@sjtu.edu.cn}}%
\thanks{$^{*}$Corresponding author: Jianping He ({\tt\small jphe@sjtu.edu.cn}).}%
}

\IEEEaftertitletext{\noindent\begin{minipage}{\textwidth}
  \centering
  \includegraphics[width=\linewidth]{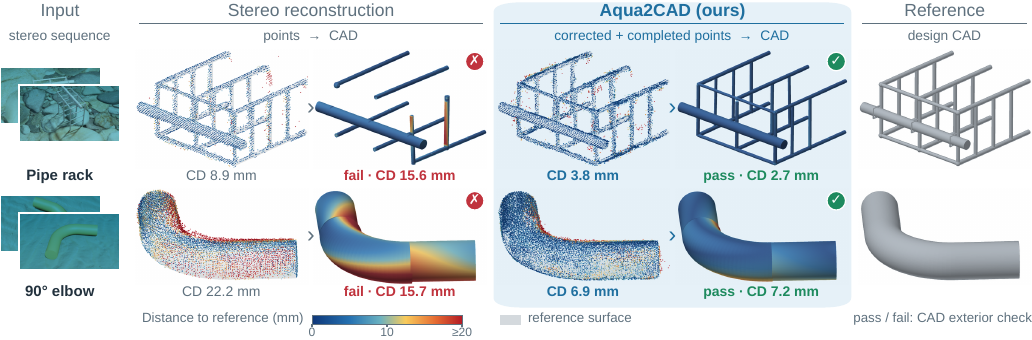}
  \teasercaption{Qualitative comparison of direct stereo reconstruction and Aqua2CAD on a pipe rack and a 90$^\circ$ elbow.
Aqua2CAD uses the original multi-view observations to correct displaced points and complete missing geometry in the original metric frame.
Both methods use the same CAD fitting backend with manually revised face groups.
Colors encode distance to the reference surface in millimeters; warmer colors indicate larger errors.
In both examples, Aqua2CAD's CAD models pass the exterior criteria, whereas the baseline fails.}
  \label{fig:aqua2cad-teaser}
\end{minipage}\par
\vspace{10pt}
}

\begin{document}
\maketitle
\thispagestyle{empty}
\pagestyle{empty}

\begin{abstract}

Recovering editable computer-aided design (CAD) models from underwater stereo images requires accurate and complete geometry, yet stereo reconstruction produces displaced points and missing regions.
The original views retain cross-view disagreement and ray evidence lost during point-cloud fusion.
Aqua2CAD exploits this evidence in two stages: view-conditioned correction and quality filtering establish reliable geometric support, while multi-view observations condition a learned shape prior to infer missing surfaces in the original metric frame.
The recovered points support surface fitting and solid assembly with manually revised face groups.
On an OceanSim-based synthetic benchmark, Aqua2CAD reduces point-cloud Chamfer distance by 60\% relative to the stereo reconstruction.
Compared with a matched point-only completion backbone, it improves missing-region recovery at 5\,mm from 16.27\% to 24.16\%.
With a shared CAD backend, exterior acceptance increases from 6/19 to 14/19.
Qualitative results on selected real underwater captures demonstrate recovery of the main structures of six 3D-printed objects.
\end{abstract}

\section{Introduction}
\label{sec:intro}

Underwater inspection and maintenance frequently require accurate measurements of existing components for replacement, adaptation, and fabrication, for example in subsea spool installation and retrofitting~\cite{chemisky2021underwater,menna2019towards}.
Such tasks benefit from computer-aided design (CAD) models with editable surfaces, explicit boundaries, and connectivity~\cite{liu2024point2cad}.
Calibrated stereo cameras, which are commonly available on underwater platforms, provide metric geometry under sufficient visibility~\cite{massotcampos2015optical,bagoren2026surfslam} and are therefore a natural source for recovering such models.

Recovering such models underwater, however, is considerably harder than on land.
Wavelength-dependent attenuation, scattering, turbidity, and changing illumination severely degrade underwater images~\cite{li2026umiunderwater,akkaynak2018revised} and, together with weak texture, impair stereo matching~\cite{bagoren2026surfslam}.
Since a practical route to CAD first reconstructs a point cloud and then fits surfaces~\cite{jobczyk2023automatic,liu2024point2cad}, the resulting errors take two forms: displaced points bias the fitted surface parameters, whereas sparse or missing regions leave surface extents and intersections poorly constrained.
Addressing these challenges requires solving two problems with different operations: correcting displaced points, which cannot recover an absent surface, and recovering missing regions, whose generated points introduce shape hypotheses that must be constrained by the available evidence.

Our key observation is that a region missing from the fused point cloud is not necessarily unobserved.
Matching, filtering, and fusion may discard geometry for which individual views still contain useful cues, and the original views retain appearance, depth disagreement, and viewing directions that the fused point set does not fully represent.
Point cloud denoising and completion methods that operate on coordinates alone~\cite{luo2021score,yu2021pointr,chen2026unico} cannot revisit this evidence, whereas image-guided completion exploits complementary image information~\cite{zhang2021vipc,xiu2025gapm}.

This observation assigns two distinct roles to the original observations.
For existing samples, disagreement across views exposes biased points, which can be corrected locally and filtered when they remain unreliable.
For missing regions, the views indicate where a surface is supported and where space is free, so they can constrain a learned shape prior wherever they remain informative, while the prior supplies hypotheses only for the truly unobserved portions.
We therefore decouple the correction of reconstructed samples from missing-shape inference and provide both stages with the original multi-view evidence: the first improves the reliability of the observed support, and the second extends its coverage by observation-conditioned generation.

We instantiate this design in Aqua2CAD (Fig.~\ref{fig:aqua2cad-teaser}).
An observation-guided local regressor aggregates view-associated image and geometric evidence and predicts bounded displacements in a geometric vector basis, and a separate quality classifier filters the corrected points.
A multi-view assisted completion model combines the shape features of the filtered points with global image tokens, spatial ray evidence, and local projected features to generate additional points in the metric frame of the reconstruction.
The resulting geometry supports analytic surface fitting and finite-solid assembly.
Since the CAD stage partitions points into face groups with ParSeNet~\cite{sharma2020parsenet} followed by manual revision, we treat CAD recovery as a human-assisted downstream test of the recovered geometry and do not claim fully automatic image-to-CAD reconstruction.

Our contributions are:
\begin{itemize}
    \item A view-associated correction mechanism that retains per-view evidence after stereo fusion and uses it for bounded geometric updates and quality filtering.
    \item A multi-view assisted completion model that conditions a learned shape prior on global image context, spatial ray evidence, and local projected features to extend the corrected points into missing regions.
    \item An evaluation on an OceanSim-based synthetic benchmark and on real underwater captures, covering geometric accuracy and coverage, component analyses, and the downstream CAD models recovered with a shared fitting backend and human-revised face grouping.
\end{itemize}

\section{Related Work}
\label{sec:related_work}

Aqua2CAD lies at the intersection of three lines of work. Underwater 3D reconstruction provides metric geometry and the original observations, point cloud refinement corrects and completes geometry, and CAD reconstruction turns geometry into editable models.

\paragraph{Underwater and multi-view 3D reconstruction.}
Underwater methods improve stereo through synthetic training and self-supervised adaptation~\cite{bagoren2026surfslam} or model the water medium in neural scene representations~\cite{levy2023seathru,li2025watersplatting,kweon2026oceansplat}.
Multi-view stereo fuses per-view depth maps after visibility and consistency checks that keep or discard samples with fixed rules~\cite{merrell2007fusion,galliani2015gipuma,schonberger2016mvs,yao2018mvsnet}, and Point-MVSNet~\cite{chen2019pointmvsnet} refines depth with multi-view point features.
These methods target dense surfaces or novel views, and the per-view evidence used during fusion is no longer available once the point cloud is formed.
We retain this evidence after fusion and reuse it to correct points and to infer missing geometry.

\paragraph{Point cloud denoising and completion.}
Learned denoisers move noisy samples toward the underlying surface using a learned score~\cite{luo2021score}, iterative filtering~\cite{de_Silva_Edirimuni_2023_CVPR}, straight-line displacement trajectories~\cite{de_Silva_Edirimuni_2024_CVPR}, structural priors~\cite{guo2026pgd}, or deterministic diffusion~\cite{liu2026dpcd}, and Noise2Score3D~\cite{wei2025noise2score3d} learns without clean targets.
Completion methods recover missing structure through geometry-aware transformers~\cite{yu2021pointr} or by jointly predicting completed points and assembly-ready primitives~\cite{chen2026unico}.
These methods operate on coordinates alone.
Image guidance has been added to both tasks: joint bilateral upsampling~\cite{kopf2007jbu} and stereo pipelines~\cite{youssefi2024cars} use color and confidence to refine depth or points, P2P-Bridge~\cite{vogel2024p2pbridgediffusionbridges3d} conditions each point on RGB and DINOv2~\cite{oquab2024dinov2} features, and ViPC~\cite{zhang2021vipc}, GAPM~\cite{xiu2025gapm}, EGIInet~\cite{xu2024egiinet}, and IAET~\cite{fang2025iaet} combine a partial cloud with a single reference image.
In these methods, image features are attached to points or drawn from one view, without the per-view depth and viewing rays of calibrated multi-view capture.
Our correction keeps the evidence tied to each viewing ray, and our completion conditions a shape prior on calibrated multi-view evidence at global, ray, and local scales.

\paragraph{CAD reconstruction from point clouds and images.}
Point-based CAD reconstruction predicts B-rep primitives and their incidence~\cite{guo2022complexgen}, fits analytic and neural surfaces and recovers edges and topology by intersection~\cite{liu2024point2cad}, or translates points into executable CAD code~\cite{rukhovich2025cadrecode}.
Image-based methods generate parametric CAD models from images~\cite{chen2025cadcrafter,kolodiazhnyi2026cadrille}, and BrepGaussian~\cite{yu2026brepgaussian} recovers B-reps from multi-view images through Gaussian splatting and labeled point sampling.
In fitting-based pipelines, sample placement determines the surface parameters, whereas coverage constrains surface extents and intersections, so their results depend on input quality that underwater stereo does not guarantee.
Rather than proposing a new CAD backend, we improve the input geometry and measure its downstream utility with a shared fitting backend.

\begin{figure*}[t]
  \centering
  \includegraphics[width=0.94\textwidth,trim=12 103 12 26,clip]{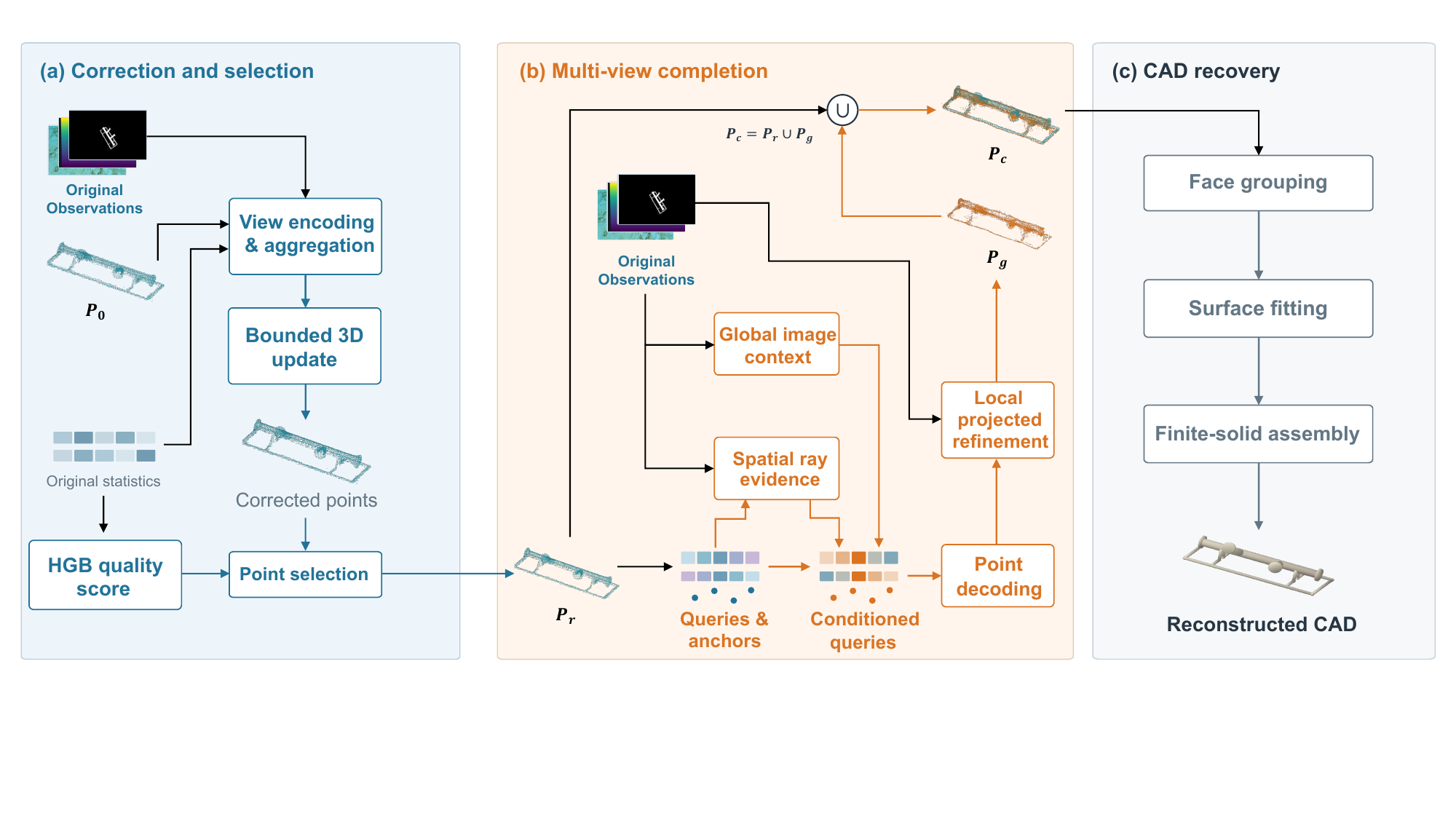}
  \caption{Overview of Aqua2CAD. The original multi-view observations guide point correction and quality filtering, and condition missing-shape inference through global image context, spatial ray evidence, and local projected refinement. Generated points are merged with the retained points in the original metric frame. Face groups from ParSeNet with manual revision drive surface fitting and finite-solid assembly.}
  \label{fig:aqua2cad-pipeline}
\end{figure*}

\section{Method}
\label{sec:method}

\subsection{Overview and View-Associated Evidence}
\label{sec:overview}
Given $N_f$ calibrated underwater stereo pairs $\mathcal I=\{(I_t^L,I_t^R)\}_{t=1}^{N_f}$, we seek metric geometry that supports CAD surface fitting and assembly (Fig.~\ref{fig:aqua2cad-pipeline}).
Stereo reconstruction yields the foreground-input reconstruction $P_0$ together with view-associated evidence $\mathcal O$, which Aqua2CAD uses in two roles:
\begin{equation}
 \begin{aligned}
 P_r&=\mathcal S\big(\mathcal R_\theta(P_0,\mathcal O)\big),\\
 P_c&=P_r\cup\mathcal N^{-1}\big(C_\phi(\mathcal N(P_r),\mathcal O)\big).
 \end{aligned}
 \label{eq:completion_frame}
\end{equation}
The regressor $\mathcal R_\theta$ corrects the existing points and the filter $\mathcal S$ removes those that remain unreliable (Sec.~\ref{sec:correction}). The completion network $C_\phi$ then generates additional points in a normalization $\mathcal N$ derived from $P_r$ (Sec.~\ref{sec:completion}).
The generated geometry extends the corrected support without replacing any retained point. The two stages are trained sequentially and share only $P_r$ and the observations.

We estimate camera motion with stereo COLMAP~\cite{schonberger2016sfm}, predict disparity with FoundationStereo~\cite{wen2025foundationstereo}, and extract the foreground with a SegFormer-based~\cite{xie2021segformer} predictor.
The resulting depth maps $D_t$, measured along the optical axis, are backprojected and fused in a common metric frame.
Standard fusion discards the per-view evidence once the point cloud is formed. We instead retain, for every fused point, its source frames, estimated cameras, RGB images, predicted depths, stereo consistency, and continuous foreground probabilities.
To compare a point with the depth predicted in each view, let $\mathbf x_i$ be the $i$-th point in world coordinates, $\mathbf T_t$ the world-to-camera transform of view $t$, $\mathbf K$ the calibrated intrinsics, and $\pi(\cdot)$ perspective division.
The projected pixel $\mathbf u_{it}$ and the signed depth residual $e_{it}$ are
\begin{equation}
 \mathbf u_{it}=\pi(\mathbf K\mathbf T_t\mathbf x_i),\qquad
 e_{it}=D_t(\mathbf u_{it})-[\mathbf T_t\mathbf x_i]_z,
 \label{eq:depth_evidence}
\end{equation}
where $[\cdot]_z$ extracts the depth coordinate in the camera frame.
A positive $e_{it}$ places the point in front of the predicted surface, a negative value behind it, and a value near zero indicates agreement.
We denote by $\mathcal V_i$ the views with a valid projection and depth for point $i$.
The residuals serve as input features for the learned updates and do not force points onto any individual depth prediction.

\subsection{Observation-Guided Correction and Selection}
\label{sec:correction}
\begin{figure}[tb]
  \centering
  \includegraphics[width=\columnwidth]{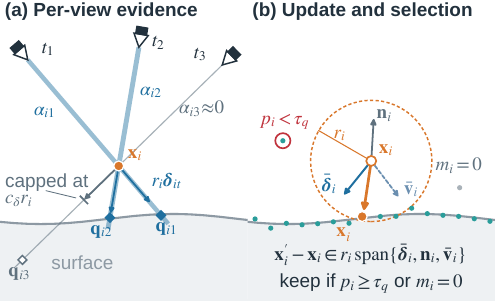}
  \caption{\textbf{Observation-guided correction and selection (schematic).} (a) A fused point $\mathbf x_i$ is compared with the points $\mathbf q_{it}$ backprojected from each view's predicted depth. Displacements are capped at $c_\delta r_i$, and line width shows the attention weight $\alpha_{it}$. (b) Bounded update and HGB selection.}
  \label{fig:local-regressor-quality-filter}
\end{figure}
\paragraph{View-associated aggregation.}
As shown in Fig.~\ref{fig:local-regressor-quality-filter}, the local neighborhood of a fused point alone cannot reveal whether the point lies on a surface or results from inconsistent depth estimates.
We therefore associate each point with up to $K$ views sampled from its source frames and encode the evidence of each view separately before aggregation.
Each view provides the foreground probability, stereo consistency, signed and absolute depth disagreement, local depth variation, and RGB patch statistics.
Let $\mathbf q_{it}$ be the world point backprojected from $D_t(\mathbf u_{it})$ and $r_i$ the local displacement radius.
The displacement toward the depth observed in view $t$, normalized by $r_i$ and capped at length $c_\delta$ without changing its direction, is
\begin{equation}
 \boldsymbol\delta_{it}=
 \frac{(\mathbf q_{it}-\mathbf x_i)/r_i}
 {\max\!\left(1,\|\mathbf q_{it}-\mathbf x_i\|_2/(c_\delta r_i)\right)},
 \label{eq:backprojection_displacement}
\end{equation}
which keeps a grossly wrong depth in one view from producing an unbounded input.
Together with the unit viewing direction $\mathbf v_{it}$ from the camera center toward $\mathbf q_{it}$ and the PCA normal $\mathbf n_i$ of the neighborhood, the geometric invariants of $\boldsymbol\delta_{it}$ and the scalar evidence form, through a shared view encoder, a view feature $\mathbf h_{it}$.
A linear layer scores each view as $\gamma_{it}=\mathbf a^{\mathsf T}\mathbf h_{it}+\beta$, and masked attention normalizes the scores over valid views only:
\begin{equation}
 \begin{gathered}
 \alpha_{it}=\frac{\mathbf 1[t\in\mathcal V_i]\exp(\gamma_{it})}{\sum_{k\in\mathcal V_i}\exp(\gamma_{ik})},\\
 \bar{\boldsymbol\delta}_i=\sum\nolimits_t\alpha_{it}\boldsymbol\delta_{it},\qquad
 \bar{\mathbf v}_i=\sum\nolimits_t\alpha_{it}\mathbf v_{it},
 \end{gathered}
 \label{eq:correction_attention}
\end{equation}
with all weights set to zero when $\mathcal V_i$ is empty.
The attention-weighted sum and the maximum of $\mathbf h_{it}$ over valid views, the point descriptor, and the valid-view fraction form the context vector $\mathbf f_i$.
Because the same weights pool the features and the geometric vectors, the views judged reliable from appearance and consistency also determine the aggregated displacement and viewing direction.

\paragraph{Bounded update in a geometric basis.}
A multilayer perceptron $\psi_\theta$, trained jointly with the view encoder and the attention layer, maps $\mathbf f_i$ to three unconstrained coefficients $(\kappa_i^{\delta},\kappa_i^{n},\kappa_i^{v})=\psi_\theta(\mathbf f_i)$, which combine the geometric vectors as
\begin{equation}
 \begin{aligned}
 \mathbf z_i&=\kappa_i^{\delta}\bar{\boldsymbol\delta}_i+\kappa_i^{n}\mathbf n_i+\kappa_i^{v}\bar{\mathbf v}_i,\\
 \mathbf x_i'&=\mathbf x_i+r_i m_i
 \frac{\mathbf z_i}{\max(1,\|\mathbf z_i\|_2)},
 \end{aligned}
 \label{eq:bounded_update}
\end{equation}
where $m_i=\mathbf 1[\mathcal V_i\neq\emptyset]$.
The update direction is thus restricted to the span of the observed displacement, the surface normal, and the viewing ray, and its length to $r_i=\operatorname{clip}(\lambda_r\ell_i,r_{\min},r_{\max})$, which scales with the local point spacing $\ell_i$. Points without valid views keep their coordinates.
The regressor is trained to predict the normalized displacement toward the nearest point on the training CAD surface, with an auxiliary proximity loss and equal total weight per object (Appendix~\ref{sec:mechanism_details}).

\paragraph{Quality filtering.}
The bounded update can relocate a locally recoverable sample, whereas a severely inconsistent point may remain unreliable after it.
After freezing the regressor, we therefore train a separate histogram gradient-boosting (HGB) classifier~\cite{ke2017lightgbm,pedregosa2011sklearn} on statistics of the original observation evidence to predict whether each corrected point lies within a tolerance $\epsilon_q$ of the training reference surface.
With its score $p_i$ and a retention threshold $\tau_q$, the filtered set is
\begin{equation}
 P_r=\{\mathbf x_i':p_i\geq\tau_q\ \text{or}\ m_i=0\}.
 \label{eq:quality_selection}
\end{equation}
The classifier inputs are computed before displacement, whereas its labels evaluate the corrected points.
Filtering preserves the coordinates and identities of the retained points, including those without valid views.

\subsection{Multi-View Completion and CAD Fitting}
\label{sec:completion}
\begin{figure}[tb]
  \centering
  \includegraphics[width=\columnwidth]{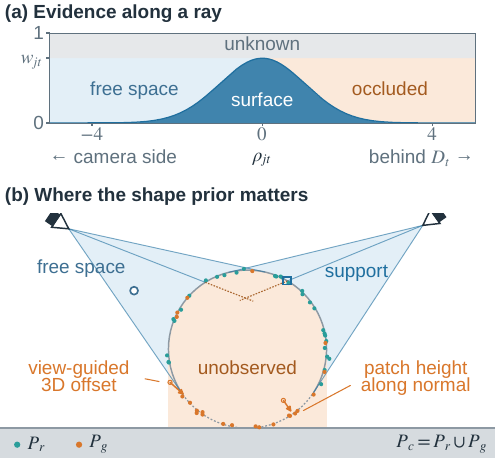}
  \caption{\textbf{Ray evidence for completion (schematic).} (a) The discrepancy $\rho_{jt}$ and the reliability $w_{jt}$ split the evidence of Eq.~\eqref{eq:ray_support} into four terms. (b) The shape prior matters most where no view gives direct support. Local refinement applies a view-guided 3D offset and then a patch-height step along estimated normals. The evidence terms are network inputs, not hard constraints.}
  \label{fig:multiview-point-cloud-completion}
\end{figure}
Moving and filtering existing points cannot introduce samples in missing regions.
Our completion model (Fig.~\ref{fig:aqua2cad-pipeline}) builds on UniCo~\cite{chen2026unico} and conditions its adaptation and decoding on the original images and camera geometry.
For $P_r$, we compute the centroid $\mathbf c$, the PCA rotation $\mathbf R$, and the bounding-box diagonal $s$, and define $\mathcal N(\mathbf x)=\mathbf R(\mathbf x-\mathbf c)/s$.
A UniCo backbone encodes $N_{\mathrm{in}}$ points sampled from $P_r$ into shape queries and coarse anchors, and $V$ views selected for directional coverage provide image features from the SegFormer encoder.
The observations then condition the shape inference at three scales, from global appearance through spatial support to the position of each generated point.
Anchors and candidates are mapped back through $\mathcal N^{-1}$ before projection, so that camera transforms and predicted depths remain metric.

\paragraph{Global and spatial conditioning.}
The sampled input points do not show what the object looks like beyond them.
Coarse image-grid tokens therefore combine appearance features with camera centers, viewing rays, predicted depth, foreground probability, and confidence, and the shape queries cross-attend to them to update their features and anchor positions with shape cues present only in the images.
Global context does not indicate where a candidate surface is supported, which ray evidence determines.
We sample spatial candidates $\mathbf y_j$ in the metric frame within the expanded bounds of the retained points and anchors and as depth hypotheses along foreground-weighted image rays.
Each candidate is projected into the selected views, where its depth discrepancy is standardized by the estimated depth uncertainty $\sigma_t$ with a floor $\sigma_{\min}$ and clipped at $\rho_{\max}$:
\begin{equation}
 \rho_{jt}=\operatorname{clip}\!\left(
 \frac{[\mathbf T_t\mathbf y_j]_z-D_t(\mathbf u_{jt})}
 {\max(\sigma_t(\mathbf u_{jt}),\sigma_{\min})},-\rho_{\max},\rho_{\max}\right).
 \label{eq:ray_evidence}
\end{equation}
With $w_{jt}$ the product of validity, foreground probability, and confidence, and $g_{jt}=\exp(-\rho_{jt}^2/2)$, each view contributes four evidence terms that sum to one (Fig.~\ref{fig:multiview-point-cloud-completion}),
\begin{equation}
 \begin{aligned}
 s_{jt}&=w_{jt}g_{jt}, & f_{jt}&=w_{jt}(1-g_{jt})\,\mathbf 1[\rho_{jt}<0],\\
 u_{jt}&=1-w_{jt}, & b_{jt}&=w_{jt}(1-g_{jt})\,\mathbf 1[\rho_{jt}>0],
 \end{aligned}
 \label{eq:ray_support}
\end{equation}
which measure surface support, free space, occlusion behind the observed surface, and unknown evidence, respectively.
A candidate in front of the observed depth lies in space that the camera saw through, whereas one behind it remains possible but unobserved, which is where the shape prior is needed.
These per-view records are encoded together with local image features before attention pooling, so that each discrepancy stays tied to its viewing ray. The shape queries attend to the resulting spatial memory, and residual branches inject its context into the query features, anchor positions, and decoder inputs.

\paragraph{Local refinement and metric output.}
Ray evidence indicates where a surface is supported but does not fix the exact position of each generated point on it.
The decoded points are therefore projected back into the $V$ views to gather local image, depth, and ray evidence, and an attention-based refiner updates their coordinates while down-weighting views that confidently place a point behind an observed surface.
A final patch-height head displaces generated points along normals estimated from nearby observed support, falling back to normals of the generated patch when this support is insufficient.
The model is initialized from the UniCo backbone, trained on the corrected and filtered inputs with the external point and image encoders frozen, and minimizes the symmetric squared Chamfer distance between the union of retained and generated points and samples of the training CAD surfaces (Appendix~\ref{sec:completion_details}).
The $N_g$ generated points are mapped back by $\mathcal N^{-1}(\mathbf q)=s\mathbf R^{\mathsf T}\mathbf q+\mathbf c$ and appended to $P_r$, preserving the metric frame without post-hoc alignment.
Although the observations condition shape inference, the generated points remain hypotheses, particularly in regions without direct support.

\paragraph{CAD fitting and finite-solid assembly.}
\label{sec:cad_fitting}
We partition the recovered point cloud into face-instance groups with ParSeNet~\cite{sharma2020parsenet} and manually revise the groups so that each represents a single surface patch. This is the only human-assisted step of the pipeline (Appendix~\ref{sec:face_revision}).
A Point2CAD-derived backend~\cite{liu2024point2cad} fits each group with the best-scoring analytic surface among planes, cylinders, cones, spheres, and tori, and estimates its finite extent from the recovered points and the multi-view observation support.
The resulting solid primitives are combined by Boolean union and cavity subtraction in Open CASCADE~\cite{occt}, ranked by point support, and refined by observation-supported member repair (Appendix~\ref{sec:cad_member_repair}).
An accepted output is a closed, positive-volume B-rep solid that survives STEP and native B-rep round trips and has explicit surface geometry and face--edge topology that standard CAD software can edit.

\section{Experiments}
\label{sec:experiments}

\subsection{Datasets and Implementation Details}
\label{sec:datasets}

\paragraph{Synthetic data.}
Our synthetic dataset, built with OceanSim~\cite{song2025oceansim}, contains 185 CAD objects in three groups of increasing geometric and structural complexity: 75 simple pipes and joints (Layer1), 75 complex assemblies such as pipe networks and support frames (Layer2), and 35 industrial parts from the ABC dataset~\cite{koch2019abc} (Layer3).
Each object is rendered against five largely featureless seabed backgrounds, such as sand and gravel, along two randomized trajectories per background, under moderate scattering combined with randomly sampled moderate or severe motion blur.
Each sequence contains 100 calibrated stereo pairs at $960\times540$ resolution, and the source CAD models provide metric reference surfaces for supervision and evaluation.
We split the objects into 148 training, 18 validation, and 19 test objects with 1,472, 180, and 190 sequences, respectively. The test split follows the group proportions with 8, 8, and 3 objects from Layer1, Layer2, and Layer3.
All learned components are trained in stages on this split.

\paragraph{Real data.}
We fabricated six objects by 3D printing their design CAD models: a 90\textdegree{} elbow, a hollow cylinder, a pipe network, a pipeline end termination (PLET), a T-joint, and a Y-joint.
The objects were placed in a water pool and recorded with the stereo camera of an underwater remotely operated vehicle (ROV) along several trajectories of 100 stereo pairs each.
The design models serve only as unregistered visual references, so the evaluation is qualitative.

\paragraph{Implementation details.}
The completion model uses $V=8$ views and generates $N_g=8{,}192$ points. Appendix~\ref{sec:experimental_setup} lists the other hyperparameters, their selection, and the training settings.

\subsection{Evaluation Metrics}
\label{sec:metrics}

\paragraph{Point clouds.}
We report the bidirectional Chamfer distance (CD, in mm) between output points and the reference CAD surface, together with the F-scores $F_5$ and $F_{20}$ at 5 and 20\,mm.
For missing-region recovery, reference samples farther than 20\,mm from the completion input form the missing set, and $R_\tau$ is the fraction of this set lying within $\tau$\,mm of the output. We report $R_5$ and $R_{20}$.
Exact definitions and the aggregation over objects are given in Appendix~\ref{sec:point_metrics}.

\paragraph{CAD models.}
An output is closed if it is a positive-volume solid that survives STEP and native B-rep round trips.
A closed output passes exterior acceptance if its mean surface distance to the reference (CAD CD), its two-sided surface coverage, its largest dimension error, and its volume error all fall within fixed tolerances relative to the reference size (Appendix~\ref{sec:cad_input_ablation}).

\subsection{Reconstruction Results and Comparisons}
\label{sec:synthetic_results}

On the 190 test sequences of 19 objects, Aqua2CAD improves both the accuracy and the coverage of the recovered geometry (Table~\ref{tab:geometry_stages}, first and fourth rows).
Relative to the foreground-input reconstruction, it reduces CD from 16.29 to 6.48\,mm (about 60\%) and raises $F_5$ from 41.75\% to 61.40\% and $F_{20}$ from 82.86\% to 94.31\%.
On the 170 sequences of 17 objects with nonempty missing regions, it achieves $R_5=24.16\%$ and $R_{20}=83.54\%$ (qualitative results and failure modes in Appendices~\ref{sec:qualitative_results} and~\ref{sec:failure_modes}).

\paragraph{Comparison with point cloud denoising methods.}
Before quality filtering, our local regressor raises $F_5$ from 43.62\% to 51.21\% relative to PGD, the strongest coordinate-only baseline on all three metrics (Table~\ref{tab:domain_denoising}).
Quality filtering further reduces CD from 14.82 to 10.95\,mm on the same subset.
Visually assisted P2P-Bridge attains the lowest CD among the external denoisers, whereas our regressor achieves a higher $F_5$. With quality filtering, our method obtains the best CD and F-scores in this comparison.
Appendix~\ref{sec:full_domain_denoising} details baseline training, checkpoint selection, and test coverage.

\begin{table}[t]
 \centering
 \caption{Denoising from identical foreground inputs, grouped into input, coordinate-only, visually assisted, and our methods. External models are fine-tuned from released weights on the 1,472 training sequences. Valid inputs count successful outputs out of 190. Metrics use the 179 sequences of 18 objects shared by all methods, excluding ten inputs unsupported by coordinate-only P2P-Bridge and one DPCD stitching failure. Both of our rows precede completion. An adapted SuperPC is reported in Appendix~\ref{sec:superpc_adaptation}.}
 \label{tab:domain_denoising}
 \footnotesize
 \setlength{\tabcolsep}{1.8pt}
 \begin{tabular*}{\columnwidth}{@{\extracolsep{\fill}}llrrrr@{}}
 \toprule
 Method & \shortstack[l]{Visual\\cues} & \shortstack[r]{Valid\\inputs} & \shortstack[r]{CD$\downarrow$\\(mm)} & \shortstack[r]{$F_5$ $\uparrow$\\(\%)} & \shortstack[r]{$F_{20}$ $\uparrow$\\(\%)} \\
 \midrule
 Foreground-input recon. & -- & 190/190 & 16.20 & 40.29 & 82.32 \\
 \midrule
 StraightPCF & -- & 190/190 & 16.41 & 43.26 & 81.77 \\
 IterativePFN & -- & 190/190 & 17.45 & 41.46 & 80.27 \\
 P2P-Bridge & -- & 180/190 & 19.52 & 37.57 & 79.75 \\
 PGD & -- & 190/190 & 15.71 & 43.62 & 82.59 \\
 Noise2Score3D & -- & 190/190 & 17.90 & 34.27 & 79.20 \\
 DPCD & -- & 189/190 & 16.69 & 43.22 & 80.93 \\
 \midrule
 P2P-Bridge & RGB+DINO & 190/190 & 11.98 & 45.92 & 83.30 \\
 \midrule
 Ours (regression) & Multi-view & 190/190 & 14.82 & 51.21 & 83.05 \\
 Ours (+ filtering) & Multi-view & 190/190 & \textbf{10.95} & \textbf{51.81} & \textbf{84.15} \\
 \bottomrule
 \end{tabular*}
\end{table}

\begin{figure}[t]
 \centering
 \includegraphics[width=\columnwidth]{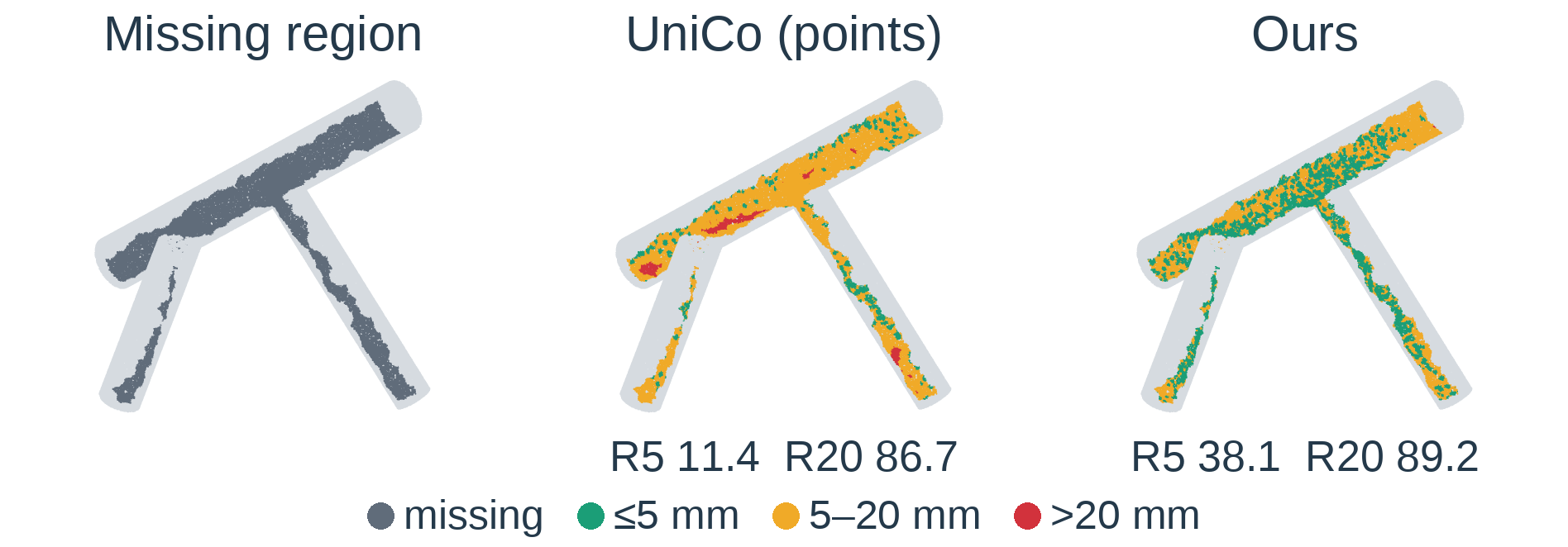}
 \caption{Missing-region recovery from identical filtered inputs on a test pipe network. Each reference sample farther than 20\,mm from the input (left) is colored by its distance to the output of point-only UniCo or of our model.}
 \label{fig:unico_r5}
\end{figure}

\paragraph{Comparison with completion methods.}
With identical filtered inputs, our multi-view assisted completion achieves the highest $F_5$ and $R_5$ among the compared methods (Table~\ref{tab:completion_methods}).
All methods are trained on the 1,472 in-domain sequences: EGIInet~\cite{xu2024egiinet} and IAET~\cite{fang2025iaet} each receive one underwater image, point-only UniCo~\cite{chen2026unico} receives coordinates only, and our model uses eight views.
Point-only UniCo and our model use the same training data and training budget, so their comparison reflects the added multi-view conditioning.
Relative to point-only UniCo, our model improves $F_5$ by 1.24 and $R_5$ by 7.89 percentage points, whereas UniCo retains a slightly lower CD and higher $F_{20}$ and $R_{20}$. Our gain therefore concentrates on recovery at the tighter 5\,mm tolerance.
Our $R_5$ is higher on 14 of the 17 objects with missing regions, and Fig.~\ref{fig:unico_r5} shows an example in which both methods cover a similar extent of the missing surface at 20\,mm, but our points lie much closer to it (see Fig.~\ref{fig:unico_r5_full} for a second example).

Both point-only UniCo and our model outperform the two single-image methods on all five metrics.
Appendix~\ref{sec:completion_baseline_protocol} gives the input and training protocols, and Appendix~\ref{sec:superpc_adaptation} reports an adapted SuperPC~\cite{du2025superpc}.

\begin{table}[t]
 \centering
 \caption{Completion from identical filtered inputs. All learned methods are trained on the 1,472 in-domain sequences and succeed on all 190 test sequences of 19 objects. CD is in mm, and F-scores and recovery rates are percentages. Recovery rates use the 170 sequences of 17 objects with nonempty missing regions. EGIInet and IAET retain the full input and append their native outputs.}
 \label{tab:completion_methods}
 \label{tab:point_only_unico}
 \footnotesize
 \setlength{\tabcolsep}{1.6pt}
 \begin{tabular*}{\columnwidth}{@{\extracolsep{\fill}}llrrrrr@{}}
 \toprule
 Method & \shortstack[l]{Visual\\cues} & CD$\downarrow$ & $F_5\uparrow$ & $F_{20}\uparrow$ & $R_5\uparrow$ & $R_{20}\uparrow$ \\
 \midrule
 Filtered input & -- & 10.53 & 52.80 & 84.99 & 0.00 & 0.00 \\
 EGIInet & 1 image & 6.86 & 55.04 & 91.21 & 7.90 & 63.64 \\
 IAET & 1 image & 6.94 & 55.35 & 92.43 & 8.66 & 72.98 \\
 UniCo & None & \textbf{6.47} & 60.16 & \textbf{94.53} & 16.27 & \textbf{83.91} \\
 Ours & 8 views & 6.48 & \textbf{61.40} & 94.31 & \textbf{24.16} & 83.54 \\
 \bottomrule
 \end{tabular*}
\end{table}

\begin{figure*}[t]
 \centering
 \includegraphics[width=\textwidth]{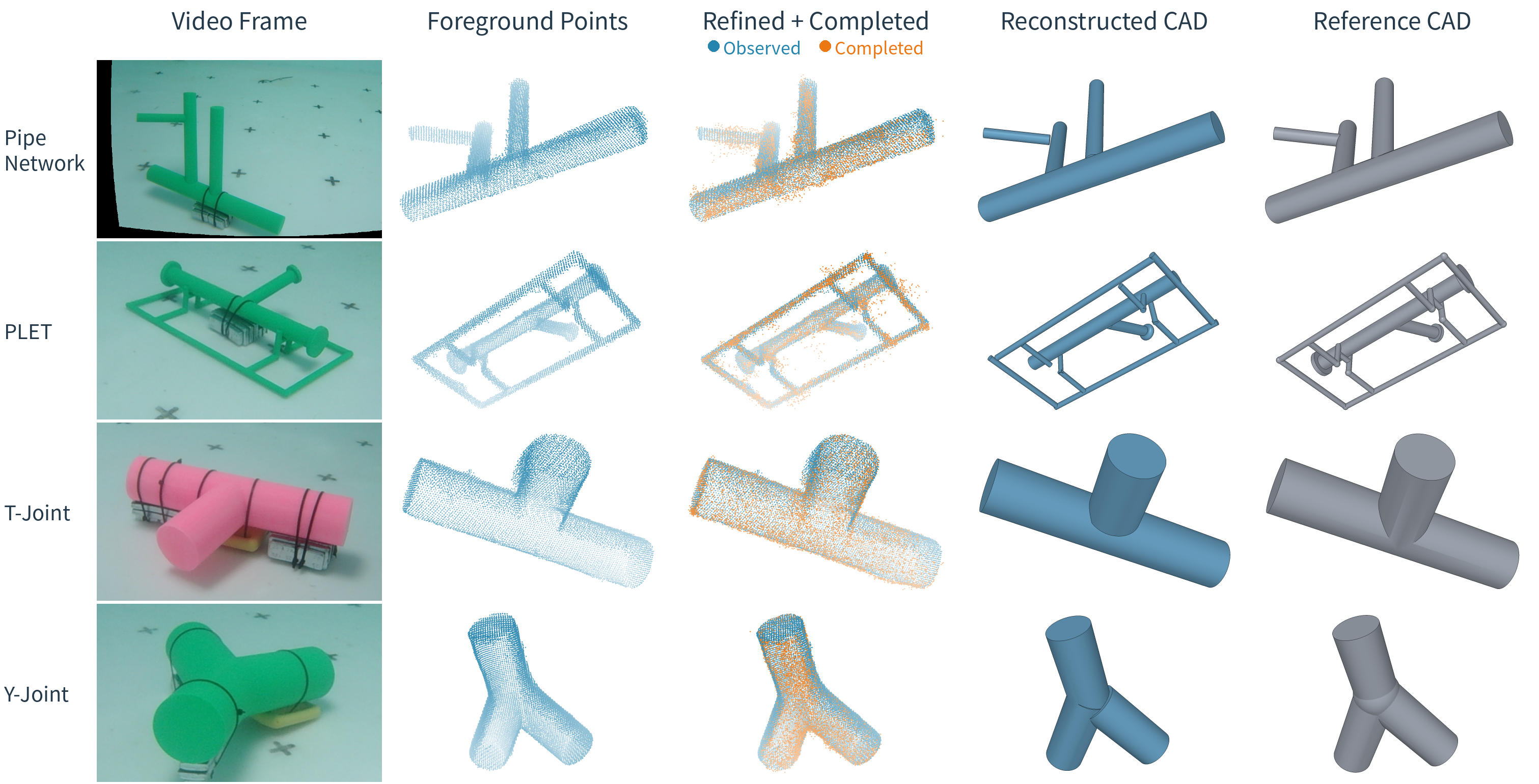}
 \caption{Real-world results for four of the six objects, one selected trajectory each (all six in Fig.~\ref{fig:real_results_all}). From left: video frame, foreground points, points after refinement and completion (blue: observed, orange: completed), reconstructed CAD, and design CAD. Columns 2--5 share one viewpoint below the object, whose underside the camera never observes. Farther points are lighter. The design model is rigidly aligned to the completed points for display only.}
 \label{fig:real_results}
\end{figure*}

\paragraph{Downstream CAD reconstruction.}
We draw one sequence per test object at random from its ten trajectories and keep this mapping fixed for the foreground-input, filtered, and completed point clouds, all of which use manually revised face groups (Sec.~\ref{sec:cad_fitting}), the same backend, and observation-supported member repair (Appendix~\ref{sec:cad_member_repair}).
Exterior acceptance rises from 6/19 to 12/19 after correction and filtering and to 14/19 after completion, while all three inputs yield 19 closed outputs and the median CAD CD decreases from 9.70 to 2.71 and then to 2.67\,mm (Table~\ref{tab:cad_reference_groups}).
Correction and filtering thus account for most of the gain in CAD accuracy, while completion adds two exterior passes.
Among the five remaining failures, the two ROV cages lack more than two thirds of their reference surface in the completion input, and completion recovers little of such large missing regions (Appendix~\ref{sec:failure_modes}).
Figure~\ref{fig:cad_atlas} shows all 19 cases, and Appendix~\ref{sec:cad_input_ablation} details the scoring protocol. A direct image-to-CAD baseline is reported in Appendix~\ref{sec:cadrille_image}.

\begin{table}[t]
 \centering
 \caption{Effect of input geometry on CAD fitting for 19 randomly sampled sequences, one per test object. All inputs use manually revised face groups and the same backend with member repair. CAD CD is averaged over the 19 outputs that are closed and scored for all three inputs.}
 \label{tab:cad_reference_groups}
 \small
 \setlength{\tabcolsep}{2.6pt}
 \begin{tabular}{lrrrr}
 \toprule
 & & & \multicolumn{2}{c}{CAD CD (mm)$\downarrow$} \\
 \cmidrule(lr){4-5}
 Input & Closed & Exterior & Mean & Median \\
 \midrule
 Foreground-input recon. & 19/19 & 6/19 & 32.63 & 9.70 \\
 Filtered & 19/19 & 12/19 & 22.62 & 2.71 \\
 Completed & 19/19 & 14/19 & 22.22 & 2.67 \\
 \bottomrule
 \end{tabular}
\end{table}

\subsection{Real-World Evaluation}
\label{sec:real_results}

We apply the regressor, quality filter, completion model, and CAD backend trained or configured on synthetic data to the real captures without fine-tuning.
Foreground masks are obtained with human assistance: an annotator corrects the masks on key frames, and SAM~2~\cite{ravi2024sam2} propagates them through each sequence.
Face groups for CAD fitting are revised manually (Appendix~\ref{sec:face_revision}), and the design models are not used at any stage.

Figure~\ref{fig:real_results} shows the reconstructed point clouds and CAD models for four objects, and Fig.~\ref{fig:real_results_all} in the appendix shows all six.
Completion appends 8,192 points to each sequence. Most coincide with observed surfaces, while the rest fill the underside, which the camera never observes because the objects rest on the pool floor or on supports.
From the completed point clouds, the reconstructed CAD models recover the main structure of each object, including the bend of the elbow, the branches of the pipe network and the joints, and the frame and header pipe of the PLET.
Errors remain, such as a step in the elbow, a crease at the Y-joint junction, and a closed hollow cylinder (Fig.~\ref{fig:real_results_all}).

\subsection{Ablation Studies and Analysis}
\label{sec:component_analysis}

\paragraph{Component contributions.}
Table~\ref{tab:geometry_stages} isolates the successive contributions of local regression, quality filtering, and completion on all 190 test sequences.
Regression improves the placement of existing samples and raises $F_5$ from 41.75\% to 52.01\%. Quality filtering further reduces CD from 14.98 to 10.53\,mm, and completion extends the retained geometry, raising $F_5$ to 61.40\% and $F_{20}$ to 94.31\%.
The pipeline thus improves the reliability of the observed support before extending its coverage.
When foreground-input, corrected, and filtered points are passed through the same frozen completion checkpoint, upstream correction and filtering raise the final $F_5$ from 50.65\% to 61.40\% (Appendix~\ref{sec:completion_input_analysis}), confirming that completion benefits from a more reliable input.
Among the regressor ablations, restricting the regressor to a single valid view per point causes the largest drop in $F_5$, 8.94 percentage points (Appendix~\ref{sec:regression_ablations_cad185}).

\paragraph{Completion components.}
To isolate the parts of the completion model, we remove one component at a time and fine-tune each variant from the same initialization as the full model, with identical data, loss, optimizer, and validation-based selection (Table~\ref{tab:completion_ablation}, Appendix~\ref{sec:completion_ablation_protocol}).
Removing local refinement causes the largest drop and halves $R_5$ from 24.16\% to 12.35\%, so generated points need view-guided adjustment to lie on the surface.
With a single view, $R_5$ falls to 16.71\%, close to the 16.27\% of point-only UniCo (Table~\ref{tab:completion_methods}), which attributes the gain in missing-region recovery to multi-view evidence.
Removing ray evidence lowers $R_5$ by 2.5 and $F_5$ by 1.8 percentage points, whereas removing global image context changes every metric by about one point or less and slightly raises $R_{20}$, so it contributes least of the four components.

\begin{table}[t]
 \centering
 \caption{Point-cloud geometry on the 190 test sequences of 19 objects. Top: each row adds one stage to the preceding row. Bottom: each completion variant removes one component and is fine-tuned once from the initialization of the full model. CD is in mm and the other metrics are percentages. Recovery rates use the 170 sequences of 17 objects with nonempty missing regions.}
 \label{tab:geometry_stages}
 \label{tab:completion_ablation}
 \footnotesize
 \setlength{\tabcolsep}{2.2pt}
 \begin{tabular*}{\columnwidth}{@{\extracolsep{\fill}}lrrrrr@{}}
 \toprule
 Setting & CD$\downarrow$ & $F_5\uparrow$ & $F_{20}\uparrow$ & $R_5\uparrow$ & $R_{20}\uparrow$ \\
 \midrule
 Foreground-input recon. & 16.29 & 41.75 & 82.86 & -- & -- \\
 + Regression & 14.98 & 52.01 & 83.55 & -- & -- \\
 + Quality filtering & 10.53 & 52.80 & 84.99 & 0.00 & 0.00 \\
 + Completion (full) & \textbf{6.48} & \textbf{61.40} & \textbf{94.31} & \textbf{24.16} & 83.54 \\
 \midrule
 \quad w/o ray evidence & 6.89 & 59.57 & 93.77 & 21.63 & 83.21 \\
 \quad w/o global image context & 6.53 & 61.18 & 94.29 & 23.02 & \textbf{84.00} \\
 \quad w/o local refinement & 7.85 & 55.76 & 91.78 & 12.35 & 77.46 \\
 \quad Single view ($V=1$) & 7.39 & 57.17 & 92.53 & 16.71 & 79.09 \\
 \bottomrule
 \end{tabular*}
\end{table}

\section{Conclusion}
\label{sec:conclusion}

We presented Aqua2CAD, which uses multi-view observations both to correct reconstructed underwater points and to condition the inference of missing geometry.
On synthetic data, it improves point accuracy and coverage, and with a common human-assisted CAD backend, exterior acceptance rises from 6/19 for the foreground-input reconstruction to 14/19. On pool captures of six 3D-printed objects, the recovered CAD models reproduce their main structure.
Face grouping remains human-assisted, and surface selection and finite-solid assembly remain bottlenecks. Automating these steps is a natural next step toward fully automatic underwater image-to-CAD~\mbox{reconstruction}.

\bibliographystyle{IEEEtran}
\bibliography{main,sec/experiment_refs}

\clearpage
\appendices
\noindent\textbf{Overview.}
Appendix~\ref{app:implementation} provides implementation details of Aqua2CAD.
Appendix~\ref{app:evaluation} defines the point-cloud and CAD evaluation protocols, and Appendix~\ref{app:baselines} describes how the baselines are adapted and evaluated.
Appendix~\ref{app:analyses} reports additional analyses and failure modes, Appendix~\ref{app:qualitative} presents qualitative results, and Appendix~\ref{app:earlier} summarizes earlier development studies.

\section{Implementation Details}
\label{app:implementation}
\label{sec:additional_experiments}

\subsection{Reconstruction and Training}
\label{sec:experimental_setup}

\paragraph{Reconstruction.}
Of the $185\times5\times2=1{,}850$ planned synthetic sequences, six training sequences failed rendering or acquisition checks and two failed SfM registration.
Camera motion is estimated from all 100 stereo pairs of a sequence with COLMAP~\cite{schonberger2016sfm}.
To reduce computation, FoundationStereo~\cite{wen2025foundationstereo} processes 50 uniformly spaced pairs at the original resolution with 32 iterations.

\paragraph{Training.}
The local regressor is trained from scratch with AdamW, a learning rate of $10^{-3}$, and a weight decay of $10^{-4}$.
Each batch contains four sequences with 256 randomly sampled points per sequence, and object-balanced weighting compensates for unequal sequence counts.
After the regressor is frozen, the HGB classifier is trained on statistics of the original observations, with labels defined by the proximity of the corrected points to the training surfaces, and a retention threshold of 0.7.
The multi-view assisted completion model is initialized from UniCo trained on the full training split and then trained on the corrected and filtered point clouds.
The three components are trained in this order, so that each downstream component learns from the outputs of the trained upstream stages.

\paragraph{Hyperparameters.}
The regressor associates each point with up to $K=16$ views and caps the normalized per-view displacement at $c_\delta=4$.
The displacement radius $r_i$ in Eq.~\eqref{eq:bounded_update} uses $\lambda_r=1.4$ and $[r_{\min},r_{\max}]=[2,12]$\,mm, which gave a higher $F_5$ than 1--6\,mm in an earlier design study (Appendix~\ref{sec:regression_design}).
The quality classifier labels corrected points within $\epsilon_q=20$\,mm of the training surface as reliable, matching the $F_{20}$ tolerance, and retains points with $\tau_q=0.7$, selected on validation data.
The completion model samples $N_{\mathrm{in}}=2{,}048$ input points, uses $V=8$ views chosen greedily to cover distinct camera directions, and generates $N_g=8{,}192$ points, 16 around each of its 512 anchors.
Ray discrepancies use $\sigma_{\min}=3$\,mm and $\rho_{\max}=8$.

\subsection{Correction and Quality Filtering}
\label{sec:mechanism_details}

The regressor uses a 16-dimensional point descriptor and 23 scalar features per view.
Four geometric invariants, $\|\boldsymbol\delta\|$, $\boldsymbol\delta^{\mathsf T}\mathbf n$, $\mathbf v^{\mathsf T}\mathbf n$, and $\boldsymbol\delta^{\mathsf T}\mathbf v$, augment the standardized view features.
A shared 27--64--64 MLP encodes each view.
Attention and maximum pooling, the point descriptor, and the valid-view fraction form a 145-dimensional input to a 64-dimensional context encoder.
The neighborhood features are fixed nearest-neighbor statistics.
For each training point, let $\mathbf p_i^*$ be its nearest point on the training CAD surface.
The normalized displacement target and the total loss are
\begin{equation}
 \begin{aligned}
 \boldsymbol\eta_i^*&=(\mathbf p_i^*-\mathbf x_i)/r_i,\\
 \mathcal L_{\mathrm{reg}}&=\mathcal L_{\mathrm{disp}}+0.25\mathcal L_{\mathrm{prox}}.
 \end{aligned}
 \label{eq:regressor_supervision}
\end{equation}
The displacement loss $\mathcal L_{\mathrm{disp}}$ averages $\|(\mathbf x_i'-\mathbf x_i)/r_i-\boldsymbol\eta_i^*\|_2^2$ over points with $\|\mathbf p_i^*-\mathbf x_i\|_2\leq0.95r_i$.
An auxiliary linear head on $\mathbf f_i$ predicts whether the original point lies within 5\,mm of the training surface, and $\mathcal L_{\mathrm{prox}}$ is its class-balanced binary cross-entropy.
Losses are weighted inversely by the number of sequences of each CAD object, so that every object receives equal total weight.
The HGB classifier uses 133 features: the 16-dimensional point descriptor, five statistics for each of the 23 original view features, and the counts of valid and source views.
Its positive labels are corrected points within 20\,mm of the training surface.
The threshold of 0.7 is selected on validation data, and no additional condition on the number of negative views is applied.

\subsection{Multi-View Assisted Completion}
\label{sec:completion_details}

\paragraph{Conditioning evidence.}
The frozen UniCo encoder provides 512 queries of dimension 384 and 512 coarse anchors.
The frozen SegFormer-B2 encoder supplies 128-channel second-stage features from the degraded RGB images.
Eight views are selected greedily to cover distinct camera directions, and each contributes a $4\times8$ global token grid, giving 256 tokens.
The spatial memory contains 512 candidates: 256 Sobol samples within the expanded bounds of the retained points and pretrained coarse anchors, and four depth hypotheses on each of eight foreground-weighted rays per view.
A valid predicted depth inside the ray--box intersection interval replaces its nearest hypothesis.
The decoder generates 16 points around each of the 512 anchors, giving 8,192 points.

The four evidence terms of Eq.~\eqref{eq:ray_support} sum to one for every candidate--view pair.
An explicit null-view option accommodates unsupported candidates.
The local refiner applies two normalized updates of the form $0.05\tanh(\cdot)$.
The final patch-height correction is additive along a support-derived PCA normal and has no hard displacement bound.

\paragraph{Training objective.}
Let $\hat P_g$ be the generated points in the normalized frame, $\hat P=\mathcal N(P_r)\cup\hat P_g$, and $G$ a set of 131,072 training reference surface samples in the same frame.
The training objective is
\begin{equation}
 \mathcal L_{\mathrm{comp}}=
 \frac{1}{|\hat P|}\sum_{\mathbf p\in\hat P}d(\mathbf p,G)^2+
 \frac{1}{|G|}\sum_{\mathbf g\in G}d(\mathbf g,\hat P)^2.
 \label{eq:completion_loss}
\end{equation}
Because the retained input is fixed during this stage, its forward-distance contribution is constant, while the reference-to-union distances account for its existing coverage.
Training uses no additional photometric or free-space loss.
The external point and image encoders are cached and frozen.
Within the completion model, the inherited feature-increase module is frozen, while the adapters, attention layers, spatial memory encoder, residual branches, decoding layers, local refiner, and patch-height head are trained with a fresh optimizer.

\section{Evaluation Protocols}
\label{app:evaluation}

\subsection{Point-Cloud Metrics}
\label{sec:point_metrics}

Let $P$ be an output point cloud, $S$ the reference CAD surface, and $Q$ a set of samples drawn from $S$.
We report the bidirectional distance
\begin{equation}
 \begin{aligned}
 \mathrm{CD}(P,S)={}&\frac{1}{2|P|}\sum_{\mathbf{p}\in P}d(\mathbf{p},S)\\
 &+\frac{1}{2|Q|}\sum_{\mathbf{q}\in Q}d(\mathbf{q},P),
 \end{aligned}
 \label{eq:physical_cd}
\end{equation}
where $d(\cdot,\cdot)$ is the Euclidean distance to the nearest surface location or point.
Precision is the fraction of output points within a tolerance of the reference surface, and recall is the fraction of reference samples within the same tolerance of the output. The F-scores $F_5$ and $F_{20}$ are their harmonic means at 5 and 20\,mm.

To evaluate the recovery of missing geometry, let $P_r$ be the filtered input of the completion stage and $P$ the completed output.
The initially missing reference samples are
\begin{equation}
 M=\left\{\mathbf q\in Q:d(\mathbf q,P_r)>20\,\mathrm{mm}\right\},
\end{equation}
where $d(\mathbf q,A)=\min_{\mathbf a\in A}\|\mathbf q-\mathbf a\|_2$.
A missing sample is recovered if the output contains a point within $\tau$\,mm of it, and for $|M|>0$ the recovery rate is
\begin{equation}
 R_\tau=\frac{\left|\left\{\mathbf q\in M:d(\mathbf q,P)\leq\tau\right\}\right|}{|M|}.
\end{equation}
The fixed 20\,mm threshold defines the missing set, whereas $\tau$ controls the required recovery accuracy.
We use $R_5$ and $R_{20}$ as the main recovery metrics.
All point-cloud metrics are averaged over the sequences of each object and then equally across objects, and recovery rates are averaged only over sequences with a nonempty missing set.

\subsection{CAD Evaluation}
\label{sec:cad_input_ablation}

Table~\ref{tab:cad_reference_groups} varies the input point clouds while keeping the sequence mapping, grouping strategy, CAD backend, and scoring protocol fixed.
One sequence per test object is drawn at random from its ten trajectories, and the mapping is fixed before the foreground-input, filtered, and completed point clouds are compared.
The fitting stage estimates surface types and parameters from the input point coordinates.
Finite solid candidates are ranked by point support, and observation-supported member repair is applied to all three inputs.
Closure follows Sec.~\ref{sec:metrics}.
Let $L$ be the bounding-box diagonal of the Boolean union of the reference solids.
A closed output passes exterior acceptance if
(i) its bidirectional mean surface distance to the reference (CAD CD) is at most $0.01L$,
(ii) at least 90\% of the predicted surface samples lie within $0.02L$ of the reference surface,
(iii) at least 90\% of the reference surface samples lie within $0.02L$ of the predicted surface,
(iv) the largest normalized dimension error over the three axes is at most 10\%, and
(v) $|V_{\mathrm{pred}}-V_{\mathrm{GT}}|/V_{\mathrm{GT}}\leq10\%$, where $V$ denotes solid volume.
CAD CD uses 32,768 area-weighted samples from each surface and exact nearest-triangle distances to the opposite surface. The same samples determine the surface fractions in criteria (ii) and (iii).
All 19 sequences remain in the closure and exterior-acceptance denominators, and CAD CD is averaged over the outputs that are closed and scored for all three inputs.
This physical-scale evaluation differs from the independently normalized and aligned evaluation of CADrille-RL in Sec.~\ref{sec:cadrille_image}.

\subsection{Human-Assisted Face Grouping}
\label{sec:face_revision}

Face grouping is the only manual step of the pipeline (Sec.~\ref{sec:cad_fitting}).
An annotator revises the groups in a browser tool that displays the recovered point cloud, colored by its ParSeNet groups, together with ten key frames of the underwater video.
The tool runs no fitting and exposes no reference geometry, reference face labels, surface types, or parameters.
The annotator merges groups that belong to the same surface and splits groups that span several surfaces by selecting points with polygon or brush tools. Selection can be restricted to the visible surface, and completed points can be hidden during selection but are never removed.
Surface types and coaxial relations can optionally be specified. When they are left empty, the fitter estimates the surface types as in the automatic setting.
Every edit is saved with undo and a revision history, and a sequence enters reconstruction only after it is marked complete and submitted as an immutable snapshot that preserves the coordinates, order, and identities of all points.
For the 17 real-world trajectories in Sec.~\ref{sec:real_results}, the annotator changed the ParSeNet grouping on 14 and confirmed it unchanged on three, yielding 109 face groups. No optional surface types or relations were entered.

\section{Baseline Protocols}
\label{app:baselines}

\subsection{Comparison Setup}
\label{sec:comparison_setup}

All point-cloud stages are evaluated on the same 190 test sequences and reference surfaces.
For the denoising comparison, StraightPCF~\cite{de_Silva_Edirimuni_2024_CVPR}, IterativePFN~\cite{de_Silva_Edirimuni_2023_CVPR}, P2P-Bridge~\cite{vogel2024p2pbridgediffusionbridges3d}, PGD~\cite{guo2026pgd}, DPCD~\cite{liu2026dpcd}, and Noise2Score3D~\cite{wei2025noise2score3d} are initialized from their released weights and fine-tuned on all 1,472 in-domain training sequences, whereas our local regressor is trained from scratch on the same sequences.
Noise2Score3D retains its unsupervised training objective.
We evaluate both coordinate-only P2P-Bridge and a visually assisted variant that conditions each point on RGB and DINOv2~\cite{oquab2024dinov2} features from eight underwater views.
Coordinate-only baselines receive point coordinates, whereas our regressor additionally uses the associated multi-view evidence. The comparison therefore controls the in-domain training data while each method retains its own inputs and initialization.
Coordinate-only P2P-Bridge requires at least 512 input points and therefore excludes ten sequences of one test object, whereas the visual variant supports all 190 sequences.
DPCD has one additional patch-stitching failure, so the matched denoising comparison uses 179 test sequences of 18 objects.
For SuperPC~\cite{du2025superpc}, we reconstruct the unreleased training components from the paper and the released code while retaining the released visual architecture and pretrained initialization. We refer to this version as SuperPC (adapted) and describe it in Sec.~\ref{sec:superpc_adaptation}.
The same SuperPC checkpoint, fine-tuned on the 1,472 filtered-input training sequences and selected on completion validation, is evaluated in both settings under the protocols of Tables~\ref{tab:domain_denoising} and~\ref{tab:completion_methods}: its native 2,048-point prediction from foreground inputs, and its native output appended to the full filtered input. Section~\ref{sec:superpc_adaptation} reports both results.
Sections~\ref{sec:full_domain_denoising} and~\ref{sec:completion_baseline_protocol} report test coverage and baseline training details.

\subsection{Denoising Baselines}
\label{sec:full_domain_denoising}

\paragraph{Coordinate-only baselines.}
StraightPCF, IterativePFN, and coordinate-only P2P-Bridge are fine-tuned on the full 1,472-sequence training split, starting from the released weights with a fresh optimizer.
Target points are the nearest projections of the training observations onto their training CAD surfaces.
Training patches contain actual stereo reconstruction errors, and no synthetic Gaussian noise is added.
When an input contains fewer points than the training patch size, observed indices are repeated together with their paired targets.
Each adaptation retains the complete released network and its losses: StraightPCF uses two velocity modules with two internal iterations, IterativePFN retains its four modules, and P2P-Bridge uses its full coordinate-based model with five inference steps.
Each epoch samples eight patches per sequence with a batch size of eight.
We train for at most 100 epochs, validate every five epochs, and stop after ten validation checks without improvement.
The learning rates are $10^{-5}$ for StraightPCF and IterativePFN and $3\times10^{-5}$ for P2P-Bridge.
The former two use Adam, whereas P2P-Bridge uses AdamW with a weight decay of $10^{-5}$, gradient clipping at 1, and mixed precision. Batches with non-finite mixed-precision outputs are replayed in full precision.

Checkpoint selection maximizes the object-averaged validation $F_5$ under a fixed reference-sample distance proxy, subject to input-relative constraints: CD at most 1.02 times, recall at most 0.01 below, and the 95th-percentile point-to-reference distance at most 1.05 times the corresponding input values.
If no checkpoint satisfies these constraints, the checkpoint with the highest validation $F_5$ is selected.
This proxy is used only for selection, and test metrics follow the exact definitions in Sec.~\ref{sec:point_metrics}.
StraightPCF reaches the 100-epoch limit and selects epoch 60.
IterativePFN and P2P-Bridge stop at epoch 50, and the fallback rule selects the released weights for both.

Pairing observations with surface targets requires three method-specific adaptations.
For StraightPCF, a zero-residual loss ratio is replaced by its continuous limit, leaving all nonzero divisions unchanged.
IterativePFN uses $n$ paired projected targets instead of its original oversampled clean target set while retaining its internal noisy intermediate targets, whose noise scale is computed from the observation-to-projection residual.
P2P-Bridge uses mean and maximum-radius normalization of the observed patch with index-paired projected targets, and retains its released online-weight inference setting.
At test time, StraightPCF and IterativePFN use patches of $\min(1000,n)$ points without padding, where $n$ is the number of input points.
P2P-Bridge uses $\min(2048,n)$ points and marks inputs with $n<512$ as unsupported. The ten such test sequences belong to one industrial part and remain in the 190-sequence coverage denominator.
The same rule yields 170 supported validation sequences for P2P-Bridge, whereas the other two methods use all 180.
Over all 190 test sequences, StraightPCF obtains 16.49\,mm CD, 44.59\% $F_5$, and 82.33\% $F_{20}$, and IterativePFN obtains 17.48\,mm, 42.91\%, and 80.92\%.

\paragraph{Visually assisted P2P-Bridge.}
The visual variant starts from the official P2P-Bridge checkpoint trained on ScanNet++~\cite{yeshwanth2023scannetpp} with RGB and DINOv2 features, using a fresh optimizer and an EMA reinitialized from the same online weights.
Eight fixed underwater left views provide three RGB channels and 384 DINOv2 ViT-S/14 channels per foreground point, giving 387 conditioning channels.
Features are gathered with the estimated cameras and observed-point visibility while preserving correspondence to the original foreground points.
Adaptation performs 100,000 updates with a batch size of four and 4,096-point patches, and all 180 validation sequences are evaluated at step zero and every 10,000 updates without early stopping.
The selection rule above chooses step 100,000, and inference uses EMA weights with ten steps.
Short inputs repeat observed indices together with their features, and stitching restores the original point identities, so all 190 test sequences yield valid outputs.
Over all 190 sequences, this variant obtains 11.65\,mm CD, 46.95\% $F_5$, and 83.87\% $F_{20}$.
Because the two variants use different fine-tuning and short-input protocols, their comparison reflects the complete variants.

\subsection{Completion Baselines}
\label{sec:completion_baseline_protocol}

Table~\ref{tab:completion_methods} uses the same filtered inputs for EGIInet, IAET, point-only UniCo, and our multi-view assisted completion model.
Point-only UniCo receives filtered coordinates without image features, retains every input point, and appends 8,192 generated points. It is trained on the same 1,472 sequences with the same training budget as our model, which follows the same retention rule and generated-point count.

EGIInet~\cite{xu2024egiinet} starts from the released eight-category checkpoint and is fine-tuned for 160 epochs, and the 180-sequence validation set selects epoch 147.
It receives the filtered point cloud and one underwater left image (frame 32 of each sequence), resized to $224\times224$ without a ground-truth crop or mask.
The estimated camera orientation and the centroid and maximum radius of the observed cloud define the coordinate transform and scale for both observations and targets.
EGIInet natively outputs 1,024 predicted points and 1,024 retained input samples. The common wrapper retains the entire filtered input and appends all 2,048 native points, including duplicates of retained samples.

IAET~\cite{fang2025iaet} starts from the official pretrained checkpoint and uses the same image, coordinate normalization, and 2,048 sampled input points as EGIInet.
We retain its four-stage architecture and losses and fine-tune for 200 epochs (4,600 updates) without early stopping, using Adam with a batch size of 64 and an initial learning rate of $10^{-3}$ that decays by a factor of 0.7 every 20 epochs.
The initial checkpoint and every epoch are evaluated on all 180 validation sequences, and the lowest object-averaged physical CD of the native output selects epoch 83.
Training and validation take 14.95 minutes on two RTX 5090 GPUs.
The native IAET output alone reaches a CD of 11.88\,mm, $F_5=21.31\%$, and $F_{20}=83.25\%$, and its recovery rates remain 8.66\% and 72.98\%.

EGIInet, IAET, and SuperPC (adapted, Sec.~\ref{sec:superpc_adaptation}) each append their 2,048 native output points to the full filtered input.
All methods complete all 190 test sequences, and the filtered input defines a nonempty missing region on 170 sequences of 17 objects.

\begin{figure*}[t]
 \centering
 \includegraphics[width=\textwidth]{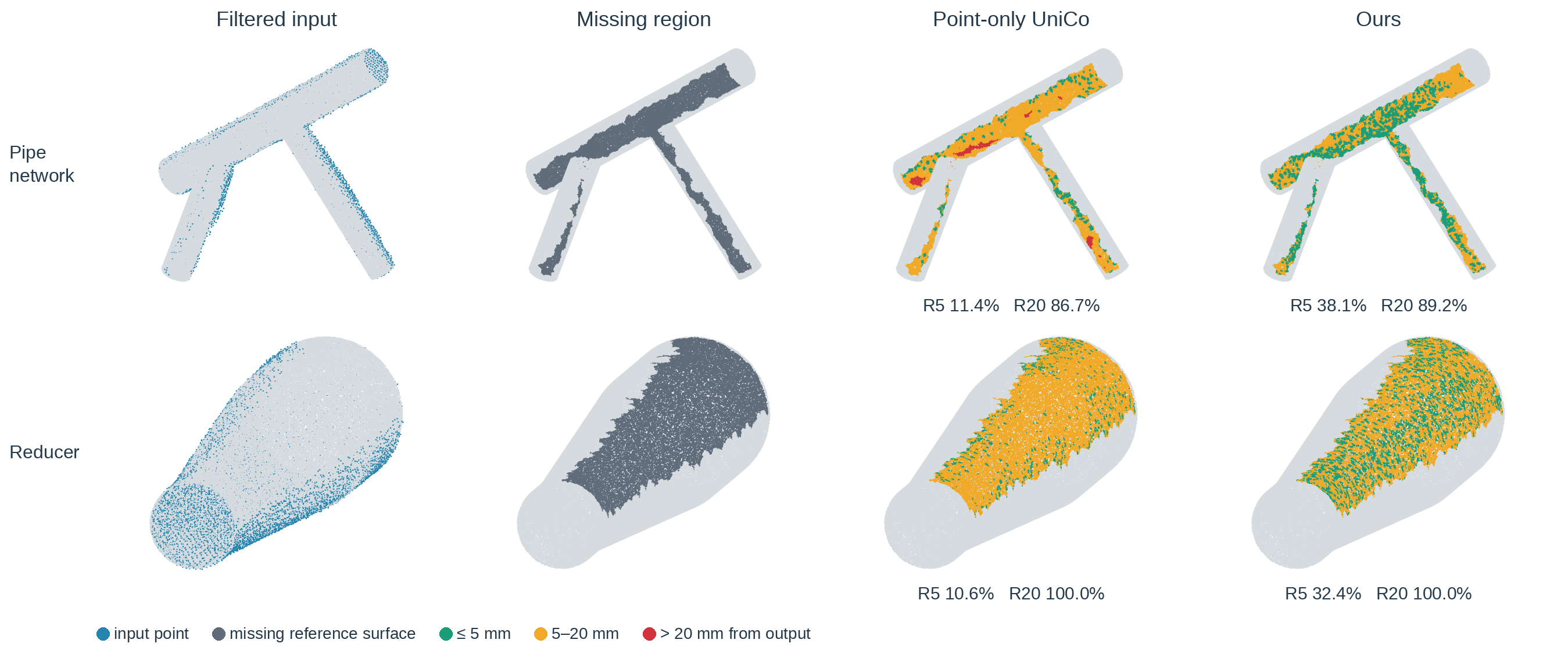}
 \caption{Missing-region recovery of point-only UniCo and our model from identical filtered inputs. Gray: reference surface samples within 20\,mm of the filtered input. Dark gray: the missing region, i.e., reference samples farther than 20\,mm from it. In the last two columns, each sample of the missing region is colored by its distance to the method output. Each row uses the viewpoint that shows the most samples of the missing region.}
 \label{fig:unico_r5_full}
\end{figure*}

Figure~\ref{fig:unico_r5_full} compares point-only UniCo and our model on two test sequences with identical filtered inputs.
In both examples, the two outputs reach a similar $R_{20}$, whereas our $R_5$ is about three times higher: 38.1\% versus 11.4\% on the pipe network and 32.4\% versus 10.6\% on the reducer.
Over the 170 test sequences with nonempty missing regions, our $R_5$ is higher on 137 sequences and on 14 of the 17 objects.
The examples were chosen to illustrate this difference and are not representative of its magnitude.

\subsection{SuperPC Adaptation}
\label{sec:superpc_adaptation}

We reconstruct the unreleased SuperPC~\cite{du2025superpc} training components from its paper and public release while preserving the released visual network and pretrained initialization.
The reconstructed objective regresses the conditional flow velocity along a linear source--target path, with interpolation time $\xi=1-\cos(\pi U/2)$ for $U\sim\mathcal{U}(0,1)$ and source jitter with a standard deviation of 0.01.
Training uses fixed exact Hungarian correspondences computed before jittering instead of online auction matching on noisy sources.
We correct the image projection to the standard pinhole form $u=f_xx/z+c_x$ and $v=f_yy/z+c_y$ after restoring the estimated camera coordinates and scaling the intrinsics to the resized image, and keep the visual encoder frozen.
The model is fine-tuned on all 1,472 training sequences with 2,048 sampled filtered points and the full underwater left image of frame 32, resized to $224\times224$ without a ground-truth crop or mask.
The centroid and radius of the observed points and the estimated cameras define the point transform.
Adam uses a global batch size of 64 on two RTX 5090 GPUs and an initial learning rate of $10^{-4}$, halved every 20 epochs.
All 180 validation sequences are evaluated every five epochs, with early stopping after ten checks without improvement and a limit of 500 epochs.
Training stops at epoch 110 after 43.08 minutes, and the lowest object-averaged physical CD selects epoch 60.
Inference uses 25 Euler steps without output clipping and produces 2,048 native points.
For completion, the output retains the entire filtered input and appends these points, following the protocol of Table~\ref{tab:completion_methods}.
For denoising, the same checkpoint receives the foreground input, sampled to 2,048 points with deterministic repetition for shorter clouds, and returns its native 2,048-point prediction, following the protocol of Table~\ref{tab:domain_denoising}. The centroid and radius are recomputed from this input with the same image and camera rules.
Neither evaluation involves additional denoising training or checkpoint selection.
On the 179 shared denoising test sequences, the native prediction obtains a CD of 39.65\,mm, $F_5=9.99\%$, and $F_{20}=58.97\%$, worse than its foreground input (16.20\,mm, 40.29\%, and 82.32\% in Table~\ref{tab:domain_denoising}).
Appended to the filtered input, it gives a CD of 9.07\,mm, $F_5=53.45\%$, $F_{20}=87.38\%$, $R_5=0.78\%$, and $R_{20}=30.21\%$ (Table~\ref{tab:completion_methods}).

\subsection{Direct Image-to-CAD Baseline}
\label{sec:cadrille_image}

We evaluate the released CADrille-RL~\cite{kolodiazhnyi2026cadrille} weights, without in-domain fine-tuning, on 19 test sequences, one per test object.
Each input consists of four underwater left images spanning the trajectory (frames 0, 32, 66, and 98).
A frozen SegFormer-B2 predicts the foreground on these images.
Each crop is centered on the predicted foreground box, all four views share one square crop size, and pixels outside the predicted foreground are set to white.
The four crops form the official $268\times268$ input grid.
Each input receives one greedy prediction capped at 768 new tokens, and the generated program is executed directly to export the resulting solid to STEP.

All 19 STEP outputs are readable, and 18 pass the readback validity and closure checks, which permit multiple solids and require valid positive-volume solids, closed oriented shells, and consistent boundary incidence.
Seventeen outputs contain two to six solids, and the output for one ROV cage has near-zero volume and fails the positive-volume check.
None of the 19 outputs passes the exterior acceptance criteria of Sec.~\ref{sec:metrics}.

For shape evaluation, each prediction and reference mesh is independently centered and normalized by its longest bounding-box side.
They are aligned by searching 24 proper axis rotations followed by rigid ICP~\cite{besl1992icp} at the normalized scale.
CD is the mean bidirectional distance from 10,000 area-weighted surface samples to the opposite triangle surface and is expressed as a percentage of the normalized longest side (Table~\ref{tab:cadrille_image}). The $F_{2\%}$ score uses 10,000 surface samples per mesh and a threshold of 0.02.

\begin{table}[t]
 \centering
 \caption{Shape accuracy of CADrille-RL from underwater images. All metrics are percentages and are computed after independent scale normalization and alignment. Shape scoring covers all 19 readable STEP outputs.}
 \label{tab:cadrille_image}
 \small
 \setlength{\tabcolsep}{3pt}
 \begin{tabular}{lrrrr}
 \toprule
 Population & $N$ & CD mean$\downarrow$ & CD median$\downarrow$ & $F_{2\%}\uparrow$ \\
 \midrule
 All outputs & 19 & 8.95 & 7.39 & 15.26 \\
 Valid closed outputs & 18 & 8.63 & 7.25 & 15.69 \\
 \bottomrule
 \end{tabular}
\end{table}

\section{Additional Analyses}
\label{app:analyses}

\subsection{Local-Regressor Ablation Protocol}
\label{sec:regression_ablations_cad185}

\begin{table}[t]
 \centering
 \caption{Local-regressor ablations on the 190 test sequences of 19 objects. All variants are trained on the 1,472 training sequences.}
 \label{tab:regression_cad185}
 \small
 \setlength{\tabcolsep}{3pt}
 \begin{tabular}{lrrr}
 \toprule
 Setting & CD (mm)$\downarrow$ & $F_5$ (\%)$\uparrow$ & $F_{20}$ (\%)$\uparrow$ \\
 \midrule
 Without RGB & 15.14 & 50.99 & 83.47 \\
 Without fine signed cues & 15.12 & 51.33 & 83.47 \\
 Single view & 16.10 & 43.07 & 82.96 \\
 Full regressor & \textbf{14.98} & \textbf{52.01} & \textbf{83.55} \\
 \bottomrule
 \end{tabular}
\end{table}

We retrain three feature variants on all 1,472 training sequences with the same 180-sequence validation set and evaluate their regression outputs on all 190 test sequences (Table~\ref{tab:regression_cad185}).
The variant without RGB removes color and RGB patch statistics.
The variant without fine signed cues removes signed depth residuals and displacement directions while retaining discrepancy magnitudes and coarse occlusion and free-space indicators.
The single-view variant retains only the first valid view per point and removes cross-view summaries and view-count features.
All variants keep the original regressor width, loss, and training schedule, and neither quality filtering nor completion is applied.
Restricting the regressor to a single view reduces $F_5$ by 8.94 percentage points, far more than removing either feature family within the multi-view model, which identifies evidence across viewpoints as the dominant factor among these ablations.
These runs use a single random initialization and do not estimate cross-seed variance.

\subsection{Completion Ablation Protocol}
\label{sec:completion_ablation_protocol}

Each variant in Table~\ref{tab:completion_ablation} starts from the same warm-start checkpoint as the full completion model and is trained on the 1,472 corrected and filtered training sequences with the same object-balanced loss, optimizer, learning-rate schedule, two-GPU batch, and seed.
Training is capped at 12 epochs, and the 180 validation sequences select the checkpoint with the same rule as for the full model, which selected epoch 9 within this cap.
The selected epochs are 9, 12, 8, and 11 for the variants in table order.
Without ray evidence, the spatial candidates, per-view ray records, ray encoder, and the residual branches that inject its context are removed.
Without global image context, the image tokens, the global cross-attention, and the image-driven anchor offset are removed.
Without local refinement, both view-guided refinement steps and the patch-height displacement are disabled.
The single-view variant keeps only the first of the eight selected views for all view-dependent inputs and draws the same number of ray candidates from that view.
Test predictions are sealed before scoring and evaluated with the same scorer as Table~\ref{tab:completion_methods}, which reproduces the full-model row exactly.
Because the shared initialization was trained with all components, the variants measure the effect of removing a component during fine-tuning rather than training without it from scratch.
Each variant is trained once, so the analysis does not estimate seed variance.

\subsection{Effect of Upstream Geometry on Completion}
\label{sec:completion_input_analysis}

To isolate the effect of upstream geometry on completion, we pass foreground-input, corrected, and filtered points through the same frozen completion checkpoint, recomputing all input features by identical rules (Table~\ref{tab:fixed_input_analysis}).
Correction raises the final $F_5$ from 50.65\% to 57.94\%, and quality filtering raises it further to 61.40\%.
Filtering also lowers CD from 10.86 to 6.48\,mm but increases $R_{20}$ by only 0.28 percentage points, with decreases on six of the 17 eligible objects.
The completion model thus responds strongly to improved upstream geometry, with a smaller and less uniform effect on missing-region recovery.

\begin{table}[t]
 \centering
 \caption{Completion with a fixed checkpoint and three upstream inputs. CD is in mm, and other metrics are percentages. CD and F-scores use 190 sequences of 19 objects, and recovery rates use 170 sequences of 17 objects with a shared nonempty missing region.}
 \label{tab:fixed_input_analysis}
 \small
 \setlength{\tabcolsep}{2.4pt}
 \begin{tabular}{lrrrrr}
 \toprule
 Input & CD$\downarrow$ & $F_5\uparrow$ & $F_{20}\uparrow$ & $R_5\uparrow$ & $R_{20}\uparrow$ \\
 \midrule
 Foreground-input recon. & 11.66 & 50.65 & 90.79 & 19.60 & 82.70 \\
 Corrected & 10.86 & 57.94 & 91.40 & 21.45 & 83.26 \\
 Filtered & \textbf{6.48} & \textbf{61.40} & \textbf{94.31} & \textbf{24.16} & \textbf{83.54} \\
 \bottomrule
 \end{tabular}
\end{table}

The checkpoint is the final completion model used in the main pipeline, trained on the 1,472 sequences of 148 objects.
The corrected input applies local regression, and the filtered input additionally applies the quality classifier.
For each input, PCA normalization, sampling to 2,048 points, eight-view selection, and the geometric and ray features are recomputed with the same rules.
The output retains all input points and adds 8,192 generated points in physical coordinates without ICP alignment.
All 570 predictions succeed, and the filtered-input outputs are identical to those of the main pipeline.
All inputs share the missing-region mask defined by distances greater than 20\,mm from the filtered input.
The 20 sequences of two objects with an empty mask remain in the CD and F-score evaluation and are excluded from the recovery rates.
Correction improves CD and $F_5$ on all 19 objects.
Filtering further improves CD on all 19 objects and $F_5$ on 18, and improves $R_{20}$ on 11 of the 17 eligible objects. Its largest decrease in recovery is 16.79 percentage points on a subsea jacket.
This analysis measures upstream-input effects for one trained completion model, and the variants are not retrained separately.

\subsection{Observation-Supported Member Repair}
\label{sec:cad_member_repair}

Member repair recovers measured components that are omitted from the assembled solid.
It identifies fitted groups that have sufficient measured support but whose point-to-surface distance increases strongly after assembly.
All nonempty replacement combinations of at most four eligible groups are considered, giving at most 15 candidates per sequence.
Candidates retain the fitted changes of the parent solid and must pass the same strict STEP and B-rep readback checks.
An observation gate requires improved coverage of the target groups, limits degradation on other supported groups and in reverse support, and requires a lower point-support cost.
The lowest-cost candidate that passes the gate is selected. Otherwise, the current solid is retained.

To fix the repair rules, we use diagnostic runs in which reference face-instance IDs supply the grouping, while surface types, parameters, boundaries, and normals are estimated from the input points.
On the 180 validation sequences, exterior acceptance increases from 93/170 to 94/170, and closure remains at 177/180. Ten sequences with invalid reference unions remain in the closure denominator.
We then apply the fixed rules unchanged to the 190 test sequences.
Of these, 159 meet the observation-sufficiency conditions and 37 trigger repair.
The search constructs 153 candidates, of which 151 pass the dual-format readback and 105 pass the observation gate, and 29 sequences adopt a changed solid.
Exterior acceptance increases from 96/190 to 98/190 without any acceptance or closure regression, and the mean CAD CD over the 189 matched closed outputs decreases from 38.27 to 36.84\,mm.
On the fixed 19-sequence subset used for the image-to-CAD comparison, closure remains at 19/19 and exterior acceptance at 10/19, while the mean CAD CD decreases from 27.80 to 27.72\,mm.
The comparison in Table~\ref{tab:cad_reference_groups} instead uses manually revised groups and applies member repair to all three inputs, and Fig.~\ref{fig:cad_atlas} shows the corresponding CAD outputs for all 19 sequences.

\subsection{Failure Modes}
\label{sec:failure_modes}

\paragraph{Large missing regions.}
Completion performs poorly when most of the surface is absent from its input.
We assign a test sequence to the large-gap group if more than 50\% of its reference surface samples lie farther than 20\,mm from its completion input.
This holds for 22 of the 190 sequences: all 20 sequences of two ROV cages, one Y-joint sequence, and one industrial part.
On these sequences, the final model recovers only 1.76\% of the missing surface within 5\,mm and 38.50\% within 20\,mm, compared with 27.18\% and 89.31\% on the remaining sequences, and its CD rises to 32.91\,mm (Table~\ref{tab:completion_gaps}).
With so little observed support, the multi-view evidence constrains only a small part of the shape, and the generated points cover the missing region coarsely and incompletely.
In the CAD comparison of Table~\ref{tab:cad_reference_groups}, both large-gap sequences, the two ROV cages, fail exterior acceptance for all three inputs.

\begin{table}[t]
 \centering
 \caption{Results of the final model grouped by the missing extent of the input. A sequence belongs to the large-gap group if more than 50\% of its reference surface samples are farther than 20\,mm from its completion input. Metrics are averaged per object within each group and then across objects. Recovery rates use the sequences with nonempty missing regions (22 of 4 objects with large gaps and 148 of 15 objects in the remainder).}
 \label{tab:completion_gaps}
 \small
 \setlength{\tabcolsep}{3pt}
 \begin{tabular}{lrrrr}
 \toprule
 Group & Seq.\ (obj.) & CD (mm)$\downarrow$ & $R_5$ (\%)$\uparrow$ & $R_{20}$ (\%)$\uparrow$ \\
 \midrule
 All & 190 (19) & 6.48 & 24.16 & 83.54 \\
 Large gaps & 22 (4) & 32.91 & 1.76 & 38.50 \\
 Remaining & 168 (17) & 4.62 & 27.18 & 89.31 \\
 \bottomrule
 \end{tabular}
\end{table}

\paragraph{CAD failures.}
Five of the 19 completed inputs still fail exterior acceptance (Table~\ref{tab:cad_reference_groups} and Fig.~\ref{fig:cad_atlas}).
Figure~\ref{fig:pipeline_pointclouds} includes the sequence with the largest gap, a Y-joint with 91.84\% of its reference samples farther than 20\,mm from the completion input, and the sequence with the largest CAD error, an ROV cage.
With manually revised face groups, these failures reflect the recovered geometry, surface selection, and finite-solid assembly rather than segmentation errors.

\section{Qualitative Results}
\label{app:qualitative}
\label{sec:qualitative_results}

\begin{figure*}[t]
 \centering
 \includegraphics[width=\textwidth]{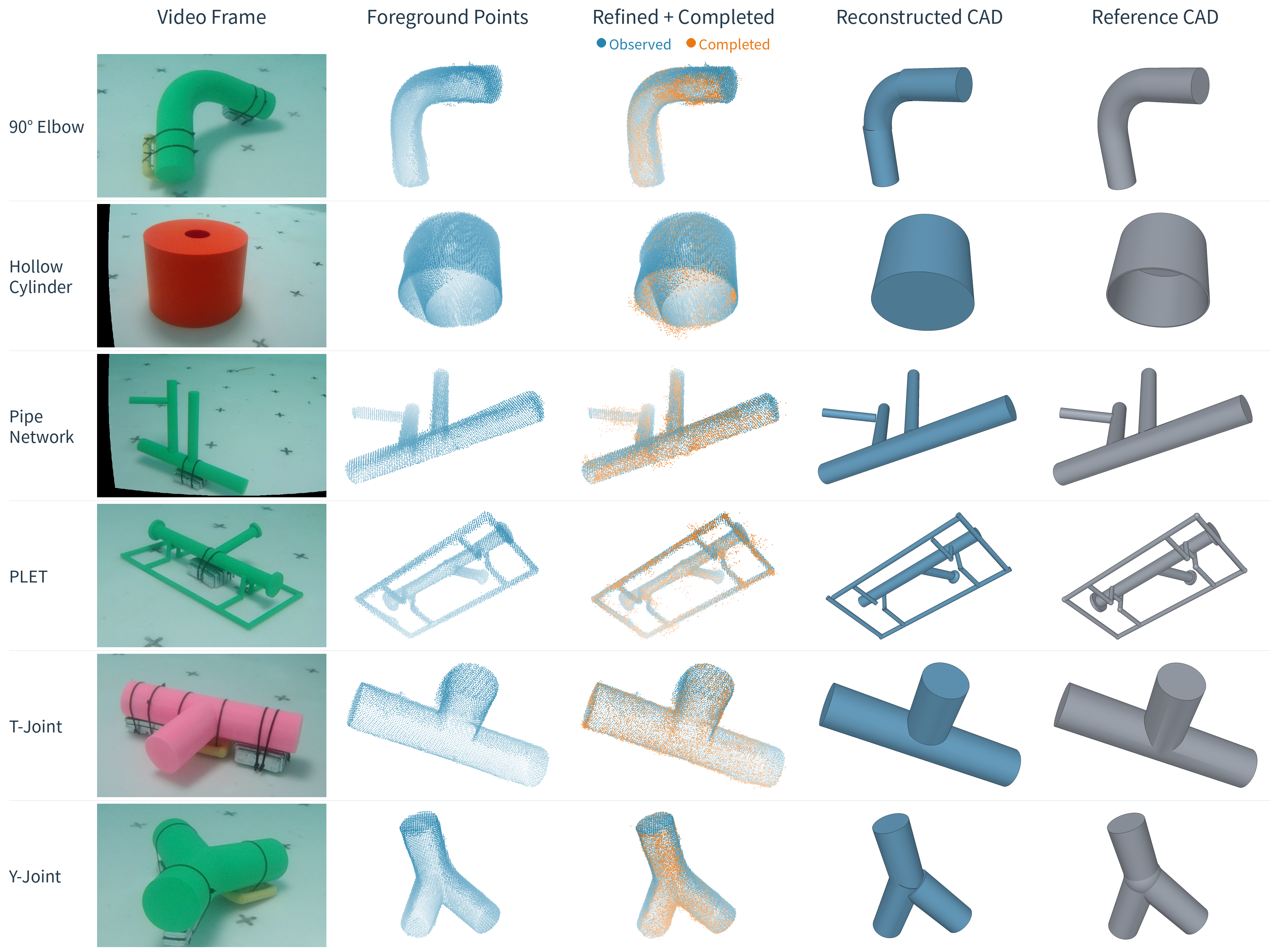}
 \caption{Real-world results for all six objects, one selected trajectory each, rendered as in Fig.~\ref{fig:real_results}. Columns 2--5 share one viewpoint per row, placed below the object. The design model is rigidly aligned to the completed point cloud for display only.}
 \label{fig:real_results_all}
\end{figure*}

Figure~\ref{fig:real_results_all} shows the real-world results for all six objects, extending Fig.~\ref{fig:real_results} with the elbow and the hollow cylinder. Their errors are discussed in Sec.~\ref{sec:real_results}.

\begin{figure*}[p]
 \centering
 \includegraphics[width=\textwidth]{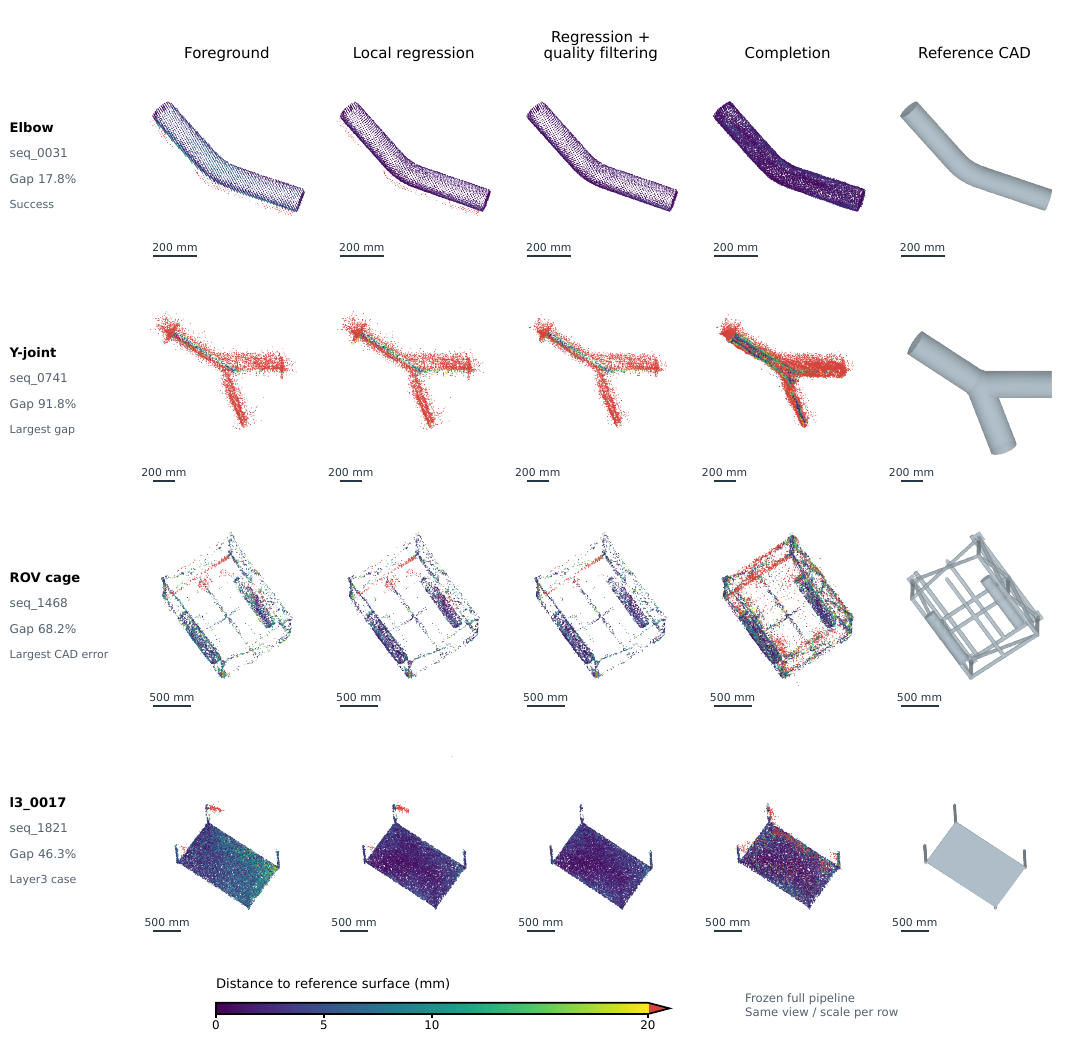}
 \caption{Successive point-cloud stages of the full pipeline. Each row uses one viewpoint and common physical bounds. Display subsampling is deterministic and does not affect the scored outputs. The gap percentage is measured with respect to the filtered completion input.}
 \label{fig:pipeline_pointclouds}
\end{figure*}

\begin{figure*}[p]
 \centering
 \includegraphics[width=\textwidth,height=0.87\textheight,keepaspectratio]{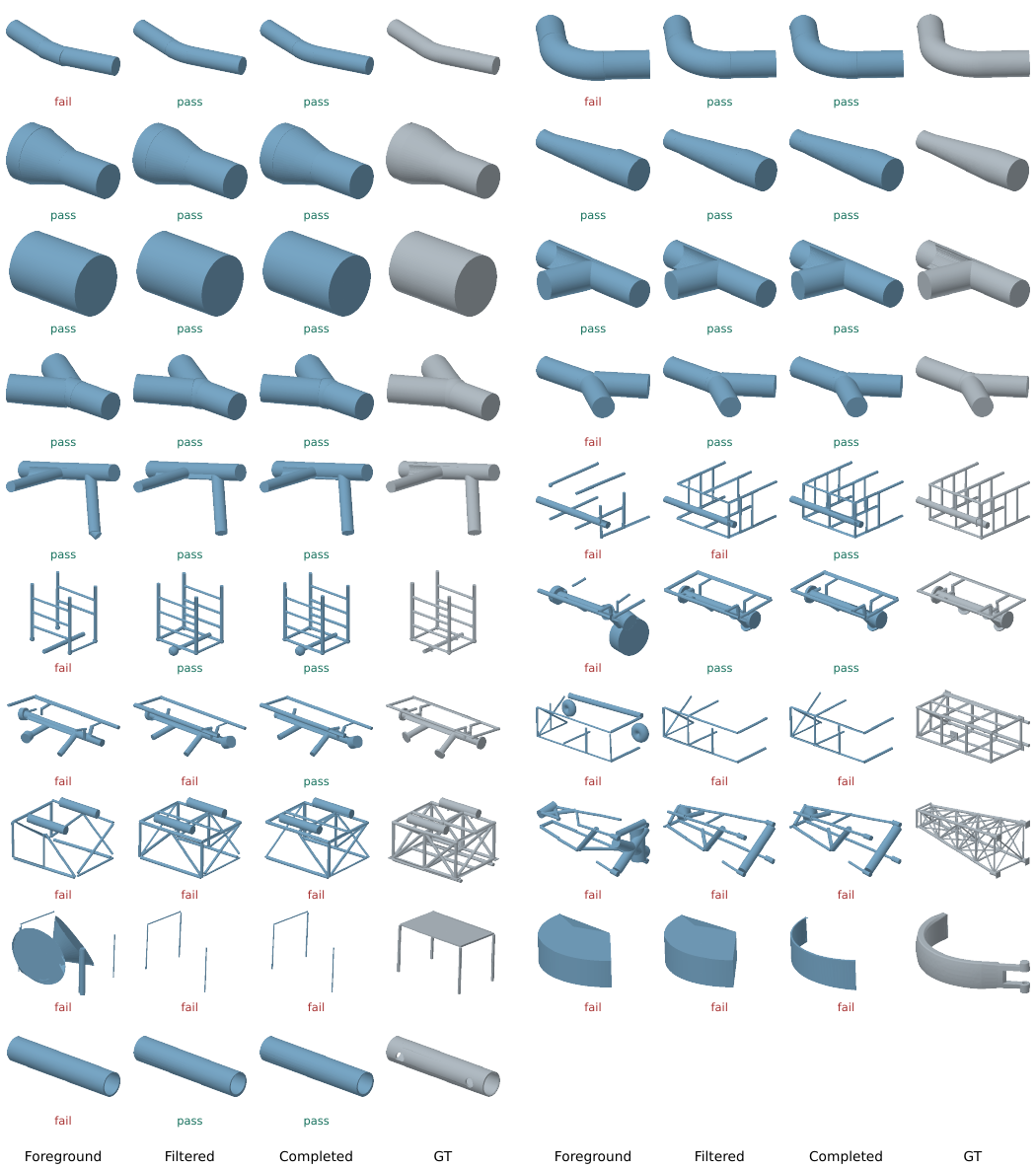}
 \caption{CAD reconstructions for all 19 inputs in Table~\ref{tab:cad_reference_groups}, one randomly sampled sequence per test object. For each object, the outputs from foreground-input, filtered, and completed points and the reference CAD (GT) are shown from left to right with the same camera and scale, retaining the full geometry. All three predictions use the same backend. Pass/fail labels follow the exterior acceptance criteria of Sec.~\ref{sec:metrics}, and GT carries no label.}
 \label{fig:cad_atlas}
\end{figure*}

Figure~\ref{fig:pipeline_pointclouds} shows four point-cloud examples.
In the example with the largest gap, 91.84\% of the reference samples are farther than 20\,mm from the filtered completion input.
Figure~\ref{fig:cad_atlas} compares the CAD outputs from foreground-input, filtered, and completed points with the reference CAD for all 19 sampled sequences in Table~\ref{tab:cad_reference_groups}.
All three inputs use the same sequence mapping and CAD backend, and the panels retain the full output extents, including excess geometry.

\section{Earlier Development Studies}
\label{app:earlier}

The studies in this section predate the current training run and use earlier data and settings. They are separate from the 190-sequence evaluation in Sec.~\ref{sec:synthetic_results}.

\subsection{Foreground Module}
\label{sec:foreground_study}

We evaluated a frozen SegFormer-B2~\cite{xie2021segformer} with high-resolution inference and local refinement on eight synthetic CAD objects, comprising 16 sequences and 800 frames.
Depth propagation, stereo consistency, and multi-view fusion improve the object-balanced IoU (Table~\ref{tab:foreground_development}).
The fusion parameters were selected on six separate development objects.
IoU improves on all eight evaluation objects, and thin-structure recall improves on seven.
On another background with the same evaluation objects, the mean IoU improves by 3.32 percentage points, whereas thin-structure recall for the three jacket objects decreases by 1.24 points.

\begin{table}[t]
 \centering
 \caption{Development evaluation of the foreground module on eight CAD objects. All metrics are percentages.}
 \label{tab:foreground_development}
 \small
 \setlength{\tabcolsep}{4pt}
 \begin{tabular}{lrrr}
 \toprule
 Setting & IoU (\%) & Precision (\%) & Thin recall (\%) \\
 \midrule
 Image predictor & 72.51 & 85.08 & 74.91 \\
 + Geometry fusion & \textbf{76.64} & \textbf{91.04} & \textbf{76.45} \\
 \bottomrule
 \end{tabular}
\end{table}

\subsection{Local-Regressor Design}
\label{sec:regression_design}

An earlier architecture study trained on 80 CAD objects and selected configurations on ten validation objects.
All variants in Table~\ref{tab:regression_development} preserve point identities and apply coordinate updates without quality filtering.
Doubling the observation-derived displacement bound from 1--6\,mm to 2--12\,mm improves $F_5$, and removing RGB statistics or fine signed depth residuals reduces it.
The latter variant retains coarse occlusion and free-space indicators and slightly improves CD, so its effect depends on the metric.
A second seed gives $F_5=46.03\%$ for the 2--12\,mm setting.
Here, CD uses fixed surface samples in both directions, whereas Sec.~\ref{sec:point_metrics} uses an exact point-to-mesh forward distance.

\begin{table}[t]
 \centering
 \caption{Regression design comparisons on the earlier ten-object validation set. CD is in mm and $F_5$ is a percentage.}
 \label{tab:regression_development}
 \small
 \setlength{\tabcolsep}{4pt}
 \begin{tabular}{lrr}
 \toprule
 Setting & $F_5$ (\%)$\uparrow$ & CD (mm)$\downarrow$ \\
 \midrule
 Foreground input & 38.88 & 17.35 \\
 Regression, 1--6 mm & 43.36 & 16.75 \\
 Regression, 2--12 mm & \textbf{46.18} & 15.94 \\
 Without RGB & 45.26 & 15.99 \\
 Without fine signed residuals & 45.20 & \textbf{15.89} \\
 \bottomrule
 \end{tabular}
\end{table}

\subsection{CAD Backend Diagnostics}
\label{sec:cad_backend_diagnostics}

A separate 19-object study used fixed earlier point clouds and automatic ParSeNet~\cite{sharma2020parsenet} groups with a Point2CAD-derived fitting backend~\cite{liu2024point2cad} and Open CASCADE~\cite{occt} assembly.
Two variants relabel completion points with uniform or observation-derived neighborhood weights, retaining the groups of observed points and requiring strong local consensus.
Table~\ref{tab:cad_development} compares these automatic settings with reference face groups.
This study uses input point clouds and settings different from those of Table~\ref{tab:cad_reference_groups}, but the same closure and exterior acceptance criteria (Sec.~\ref{sec:metrics}).
Correct grouping alone does not resolve all fitting and assembly errors.
The study involves no manual editing and precedes the current point-cloud pipeline.

\begin{table}[t]
 \centering
 \caption{Grouping diagnostics on fixed earlier inputs for 19 CAD objects. Closure and exterior acceptance follow Sec.~\ref{sec:metrics}.}
 \label{tab:cad_development}
 \small
 \setlength{\tabcolsep}{4pt}
 \begin{tabular}{lrr}
 \toprule
 Grouping & Closed & Exterior pass \\
 \midrule
 ParSeNet & 17/19 & 3/19 \\
 Uniform propagation & 17/19 & 3/19 \\
 Observation-weighted & 17/19 & 3/19 \\
 Reference face IDs & 19/19 & 6/19 \\
 \bottomrule
 \end{tabular}
\end{table}

We also evaluated BrepGaussian~\cite{yu2026brepgaussian} on six objects, two per shape group.
Each object uses 50 underwater left images, stereo-SfM camera poses, 30,000 geometry-and-edge optimization steps, and 15,000 segmentation optimization steps.
Masks receive no manual correction, and the surface fitter exports solids through Open CASCADE.
One of the six objects yields a valid closed solid, and none passes the complete acceptance checks of that study, which include observed-input coverage and volume consistency after STEP round trips.
This automatic six-object study does not provide a matched comparison with the current human-assisted pipeline.

\end{document}